\documentclass{article}

\usepackage{PRIMEarxiv}
\usepackage{natbib}
\usepackage[utf8]{inputenc} 
\usepackage[T1]{fontenc}    
\usepackage{array}
\usepackage{multirow}
\usepackage{hyperref}       
\usepackage{url}            
\usepackage{booktabs}       
\usepackage{amsfonts}       
\usepackage{nicefrac}       
\usepackage{microtype}      
\usepackage{lipsum}
\usepackage{fancyhdr}       
\usepackage{graphicx}       
\usepackage{amsmath}
\usepackage{lineno}
\usepackage{xcolor}

\graphicspath{{media/}}     

\newcommand{\best}[1]{\textcolor{purple}{\mathbf{#1}}}
\newcommand{\sbest}[1]{\textcolor{brown}{\mathbf{#1}}}

 \usepackage{xspace}

\newcommand{\alg}{\textsc{\small{GmFlow}}\xspace}

\newtheorem{proof*}{Proof}[section]

\title{Large-Scale 3D Ground-Motion Synthesis with Physics-Inspired Latent Operator Flow Matching
}

\author{
Yaozhong Shi$^{1,*}$ \And
Grigorios Lavrentiadis$^{2}$ \And
Konstantinos Tsalouchidis$^{3}$ \And
Zachary E. Ross$^{4}$ \And
David McCallen$^{3}$ \And
Caifeng Zou$^{4}$ \And
Kamyar Azizzadenesheli$^{5}$ \And
Domniki Asimaki$^{1,*}$
}

\usepackage{xcolor}
\usepackage{todonotes}

\begin{document}
\maketitle

\begingroup
\renewcommand\thefootnote{\fnsymbol{footnote}}
\footnotetext[1]{%
$^{1}$Division of Engineering and Applied Sciences, California Institute of Technology, Pasadena, CA, USA;
$^{2}$School of Engineering and Applied Sciences, University at Buffalo, Buffalo, NY, USA;
$^{3}$Lawrence Berkeley National Laboratory, Berkeley, CA, USA;
$^{4}$Seismological Laboratory, California Institute of Technology, Pasadena, CA, USA;
$^{5}$NVIDIA, Santa Clara, CA, USA.
Corresponding authors: Yaozhong Shi (\texttt{yshi5@caltech.edu}) and Domniki Asimaki (\texttt{domniki@caltech.edu}).}
\endgroup

\begin{abstract}
Earthquake hazard analysis and design of spatially distributed infrastructure, such as power grids and energy pipeline networks, require scenario-specific ground-motion time histories with realistic frequency content and spatiotemporal coherence. However, producing the large ensembles needed for uncertainty quantification with physics-based simulations is computationally intensive and impractical for engineering workflows. To address this challenge, we introduce Ground-Motion Flow (\alg), a physics-inspired latent operator flow matching framework that generates realistic, large-scale regional ground-motion time-histories conditioned on physical parameters. Validated on simulated earthquake scenarios in the San Francisco Bay Area, \alg generates spatially coherent ground motion across more than 9 million grid points in seconds, achieving a 10,000-fold speedup over the simulation workflow, which opens a path toward rapid and uncertainty-aware hazard assessment for distributed infrastructure. More broadly, \alg advances mesh-agnostic functional generative modeling and could potentially be extended to the synthesis of large-scale spatiotemporal physical fields in diverse scientific domains.

\end{abstract}
\section{Introduction}

\begin{figure*}[ht]
    \vspace*{-.3cm}
    \centering

    \includegraphics[width=0.95\textwidth,trim={1cm, 0, 1cm, 0},clip]
    {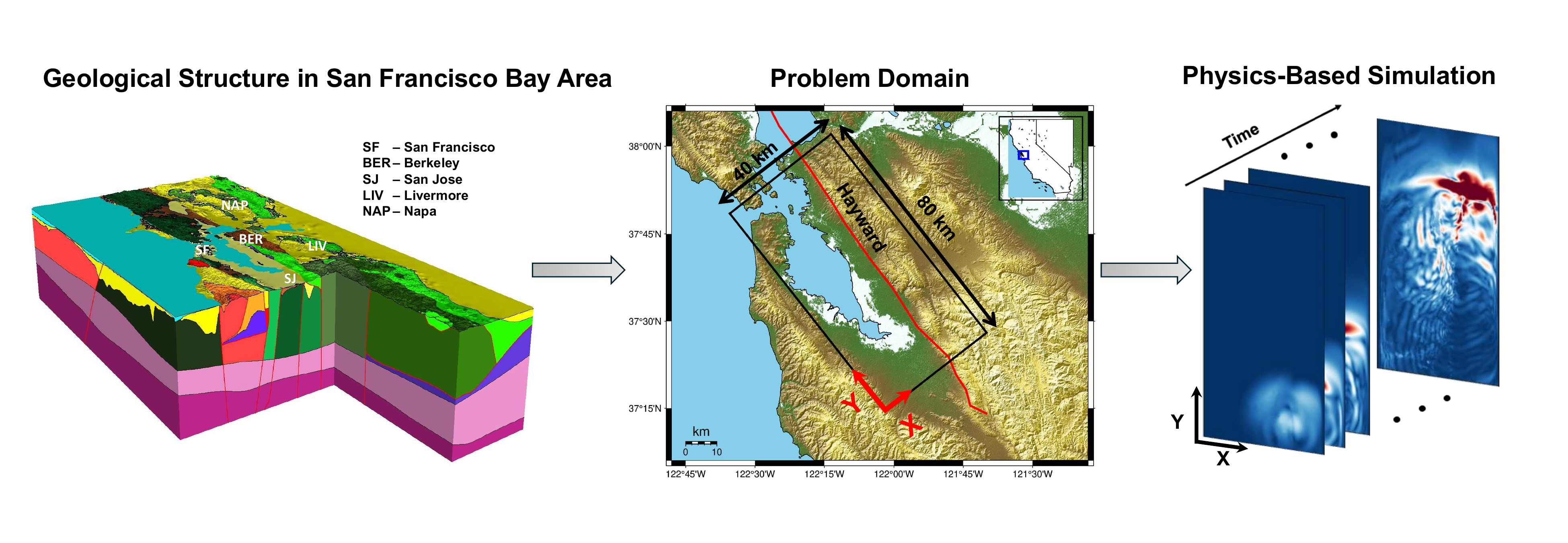}
\caption{\textbf{Workflow for physics-based ground-motion simulation in the San Francisco Bay Area.}
  \textbf{(left)} A 3D representation of the heterogeneous geological structure beneath the Bay Area, adopted from https://www.usgs.gov/media/ images/3d-geologic-model with permission.
  \textbf{(center)} The problem (computational) domain used for the numerical simulations.
  \textbf{(right)} An example of finite-fault event simulation.}
  \label{fig:data_demo}   
\end{figure*}

Managing and operating modern infrastructure systems, such as electric power grids, nuclear power plants, and other critical lifeline networks, requires rigorous evaluation of the performance and risk of systems composed of multiple, interdependent components~\cite{guidotti_modeling_2016}.
In seismic-prone regions, designing these assets for resilient performance under strong shaking is crucial, as their failure can trigger cascading disruptions, endanger lives, and impose major economic losses. 
Consequently, the design process requires scenario-based hazard characterization and the quantification of within-event and between-event uncertainties to calculate the joint exceedance of performance thresholds with ground-motion time histories~\cite{miller_groundmotion_2015,baker_seismic_2021}. 
In principle, event-level regional ground-motion time histories can be obtained either from empirical ground-motion databases or from physics-based simulations (Fig~\ref{fig:data_demo}). In practice, however, both approaches face fundamental limitations. Observational records of large-magnitude earthquakes remain sparse and unevenly distributed, and the application of scenario-specific filtering criteria (e.g., magnitude, distance, mechanism, site conditions) further reduces the number of usable time histories. Physics-based rupture and wave-propagation simulations partially alleviate data scarcity by enabling direct modeling of scenarios of interest at prescribed spatial resolution. Nevertheless, solving the elastodynamic wave equation in heterogeneous media to generate broadband ground motions at regional scale remains computationally intensive. Moreover, the resolution and fidelity of simulated ground motions are inherently constrained by the quality and resolution of the input models (e.g., three-dimensional shear-wave velocity structure), which often lack sufficient detail to accurately propagate high-frequency content critical for civil infrastructure applications~\cite{rodgers_regionalscale_2020, pitarka_broadband_2013, lavrentiadis_seismologically_2025}. 
These limitations motivate a new paradigm for generating scenario-consistent broadband waveforms, namely machine-learning-based surrogate approaches capable of rapidly producing ensembles of scenario-conditioned, spatially coherent ground-motion time histories while preserving key physical characteristics.

Recent studies have explored conditional generative models capable of synthesizing seismic waveforms across broadband frequencies at specific sites ~\cite{florez_datadriven_2022,shi_broadband_2024, palgunadi_high_2025}. Although promising, these approaches typically treat ground motions as independent one-dimensional time series at isolated sites, thereby failing to capture the within-event spatial correlations necessary for analyzing distributed systems; Extending the generative framework to synthesize coherent regional wavefields over large and continuous spatial domains requires a fundamental shift in problem formulation, introducing significant challenges in learning algorithm design, model architecture, and computational scalability during both training and inference, all of which must be addressed to yield a tractable and physically consistent solution.

\begin{figure*}[ht]
    \vspace*{-.3cm}
    \centering

    \includegraphics[width=0.95\textwidth,trim={1cm, 0, 1cm, 0},clip]
    {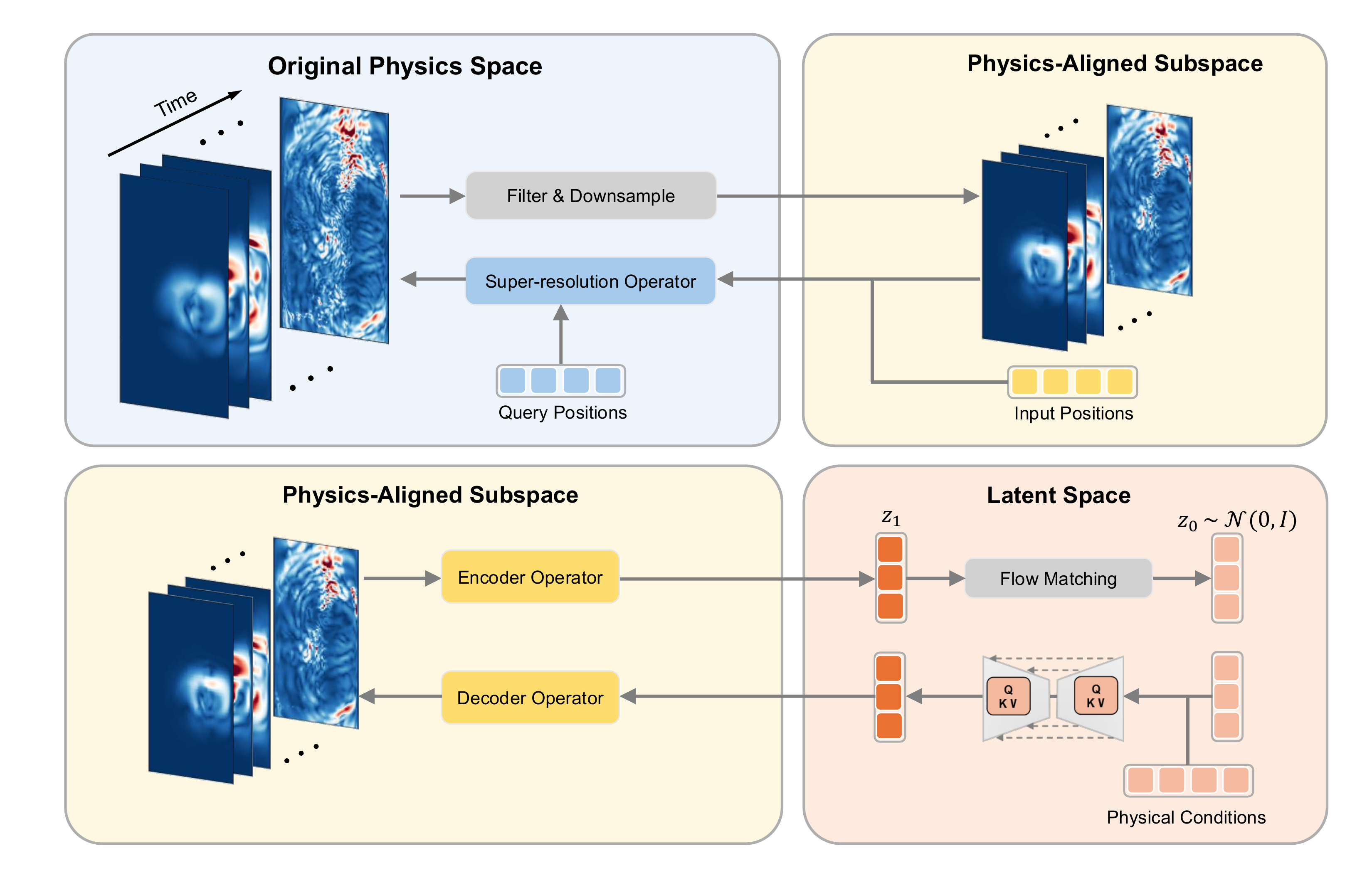}
\caption{\textbf{\alg model architecture and two-stage synthesis pipeline.}
\textbf{(top)} The original surface wavefield $u$ is filtered and downsampled to a low-frequency representation $u_f$; the reverse mapping is performed by a super-resolution neural operator.
\textbf{(bottom)} An encoder--decoder neural operator encodes $u_f$ to a latent code $z_1$, and decodes it back to $u_f$; conditional flow matching transports $z_0\!\sim\!\mathcal{N}(0,I)$ to $z_1$ given physical conditions $c$.}

 \label{fig:demo}   
\end{figure*}

To address these challenges, we propose Ground-Motion Flow (\alg), building on the machine-learning framework introduced in \cite{liu_flow_2022, rombach_high-resolution_2022, shi_stochastic_2025}. To the best of our knowledge, \alg is the first large-scale 3D ground-motion generative model formulated directly in function space. The method leverages neural operators \cite{li_fourier_2021} and follows a simple but effective two-stage design (Fig.~\ref{fig:demo}): it first generates a low-frequency regional wavefield that captures the dominant large-scale spatiotemporal structure and coherence of earthquake shaking, and then reconstructs a full-band wavefield at the target resolution from that low-frequency realization. In the first stage, an autoencoding operator maps the low-frequency wavefield into a compact latent representation, and a conditional flow-matching model efficiently samples from this latent space to generate spatially coherent regional realizations. In the second stage, a discretization-agnostic super-resolution neural operator reconstructs the full-band wavefield conditioned on the low-frequency field, recovering higher-frequency effects associated with scattering, heterogeneity, and localized amplification. Together, these components form an end-to-end resolution-agnostic generative pipeline that bypasses the computational bottleneck of direct high-dimensional generation while still producing physically consistent wavefields across different scales. We demonstrate that \alg achieves unprecedented computational scalability,  providing approximately a 10{,}000-fold acceleration in inference relative to physics-based simulations, enabling the training of full regional 3D ground-motion generative models for the San Francisco Bay Area on a single workstation GPU, while also appropriately capturing the complex multiscale physics of large-magnitude earthquakes; this enables large ensembles of scenario-consistent ground motions, which are important for uncertainty quantification and infrastructure risk assessment.

A central feature of \alg is the introduction of a \textit{Physics-Aligned Subspace}, in which the intermediate generation step is performed in a physically meaningful low-frequency representation. This design greatly reduces the computational cost of training and inference. In addition, because \alg is formulated as a functional generative model, it supports zero-shot super-resolution, allowing the generated wavefields to be evaluated at spatial resolutions finer than those used during training. Such resolution flexibility is especially important in scientific machine learning, where predictions often need to be queried on refined grids for downstream analysis. More broadly, we envision \alg serving as a foundational framework for the generative modeling of diverse large-scale spatiotemporal natural-hazard phenomena, including hurricanes, wildfires, tsunamis, and volcanic eruptions~\cite{schwerdtner_uncertainty_2024,yu_probabilistic_2025, price_probabilistic_2025}, thereby advancing uncertainty quantification, risk mitigation, and resilient infrastructure planning.

\begin{figure*}[ht]
    \vspace*{-.3cm}
    \centering

    \includegraphics[width=0.95\textwidth,trim={1cm, 0, 1cm, 0},clip]
    {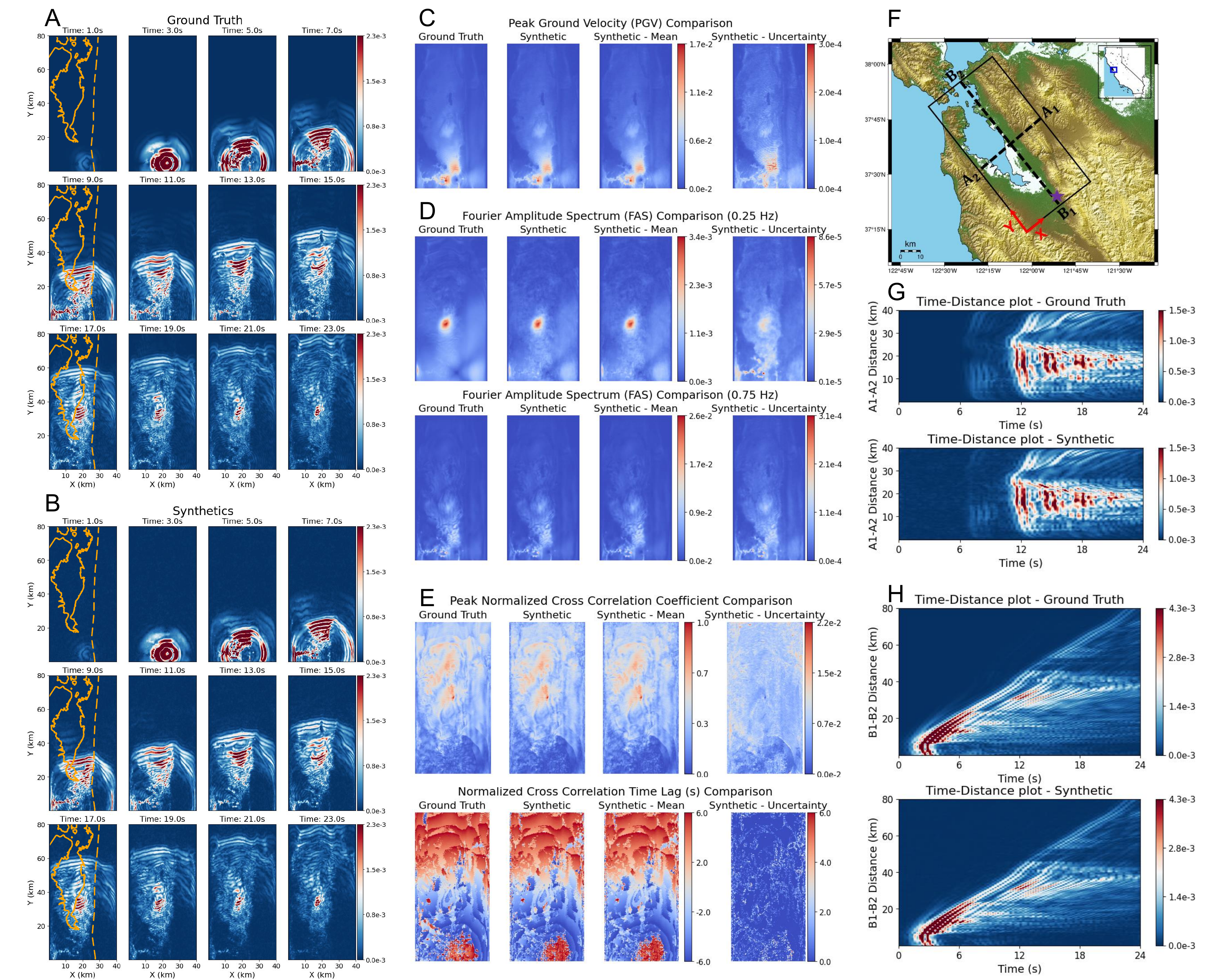}
 \caption{\textbf{Evaluation of \alg on a representative $M_w\,4.4$ point-source scenario.}
\textbf{(A--B)} Side-by-side comparison of the ground-truth velocity wavefield and a single stochastic realization generated by \alg at selected time steps (snapshots). Orange dash and solid lines represent Hayward fault and coastline, respectively.
\textbf{(C)} Peak ground velocity (PGV; $m/s$), comparing the ground truth, one synthetic realization, the ensemble mean ($N=100$), and the predictive uncertainty.
\textbf{(D)} Spatial maps of the Fourier amplitude spectrum (FAS) at 0.25 Hz (top) and 0.75 Hz (bottom).
\textbf{(E)} Normalized cross-correlation (NCC) for spatiotemporal coherence analysis, showing the peak correlation coefficient (top) and the corresponding time lag (bottom).
\textbf{(F)} Geographic map of the computational domain, with epicenter (purple star) and the virtual seismic profiles A1--A2 and B1--B2 indicated.
\textbf{(G--H)} Time--distance plots along profiles A1--A2 and B1--B2.}\label{fig:point_source}   
\end{figure*}

\begin{figure*}[ht]
    \vspace*{-.3cm}
    \centering

    \includegraphics[width=0.95\textwidth,trim={1cm, 0, 1cm, 0},clip]
    {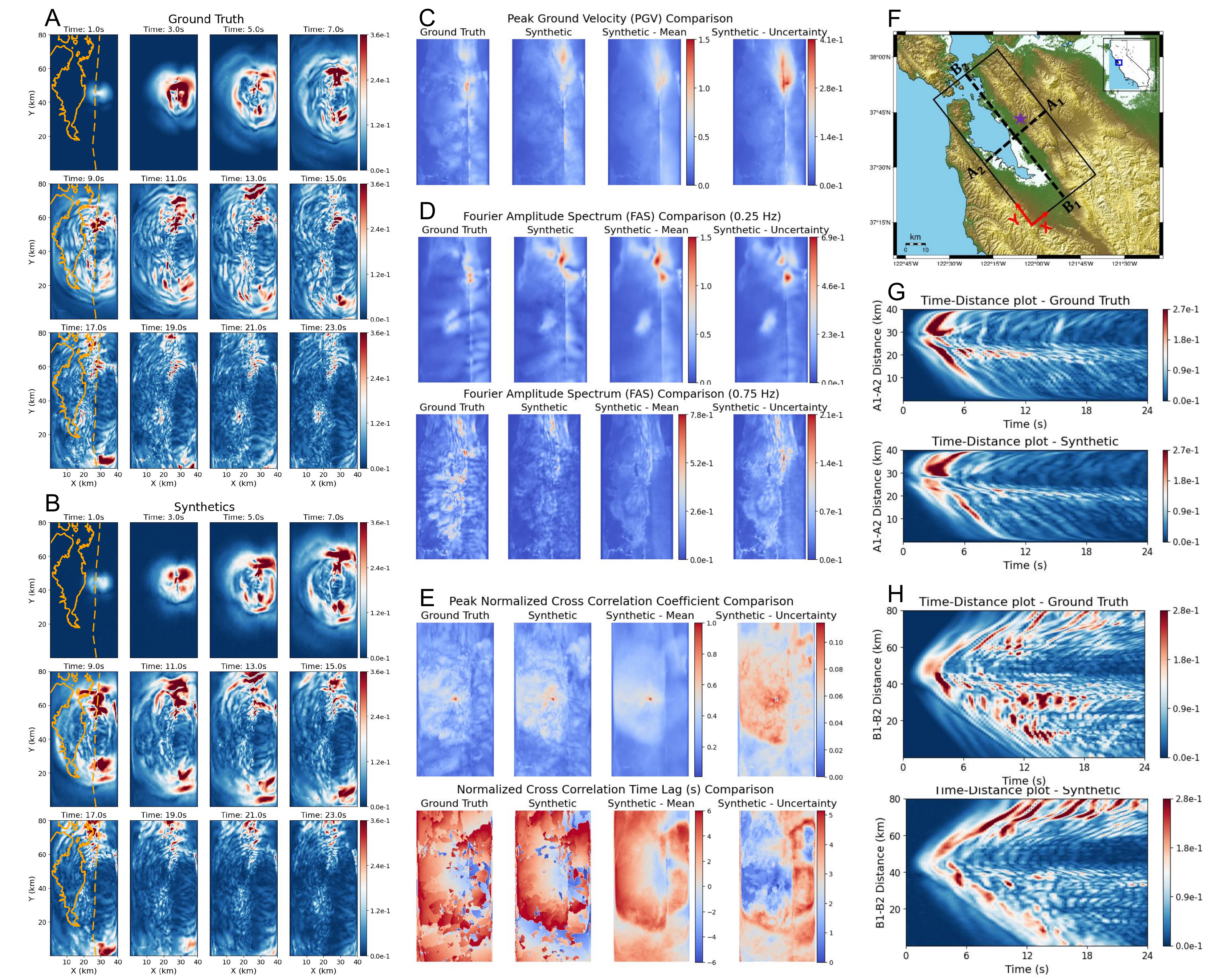}
     \caption{\textbf{Evaluation of \alg on a representative $M_w\,7.0$ fintie-rupture scenario.} }\label{fig:extended_source}   
\end{figure*}

\begin{figure*}[ht]
    \vspace*{-.3cm}
    \centering

    \includegraphics[width=0.95\textwidth,trim={1cm, 0, 1cm, 0},clip]
    {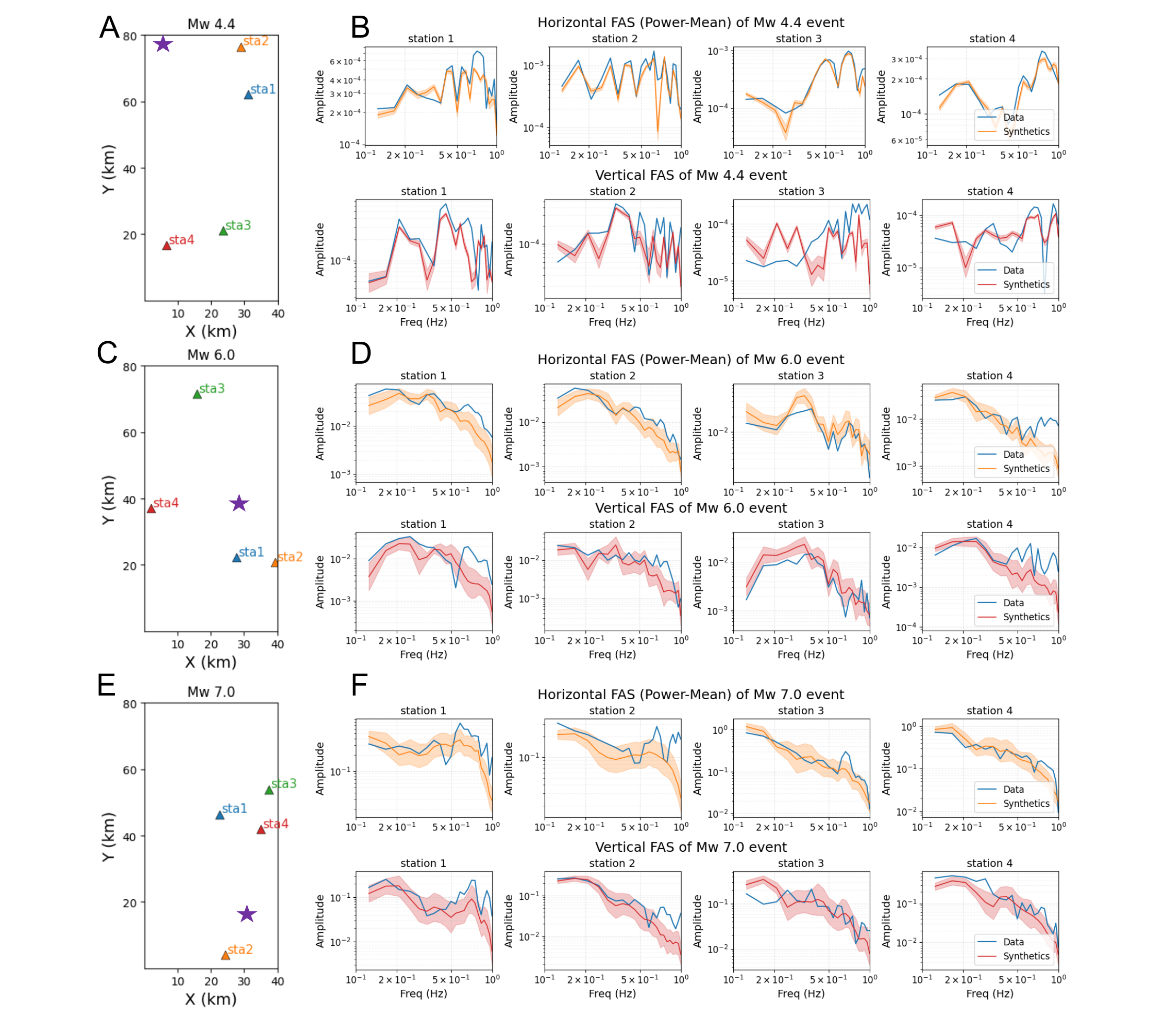}
    \caption{\textbf{Station-level spectral evaluation across magnitude range.}
    \textbf{(A, C, E)} Spatial distribution of four randomly selected validation stations (triangles) and the epicenter (star) for $M_w$ 4.4, 6.0, and 7.0 scenarios. \textbf{(B, D, F)} Comparison of the horizontal (power-mean) and vertical Fourier Amplitude Spectra (FAS) for the ground truth (blue) and synthetic realizations (orange/red). The solid lines for the synthetics represent the geometric mean of $N=100$ realizations, while the shaded regions indicate the $\pm1$ geometric standard deviation.
    } \label{fig:site_spectra}   
\end{figure*}

\begin{figure*}[ht]
    \vspace*{-.3cm}
    \centering

    \includegraphics[width=0.95\textwidth,trim={1cm, 0, 1cm, 0},clip]
    {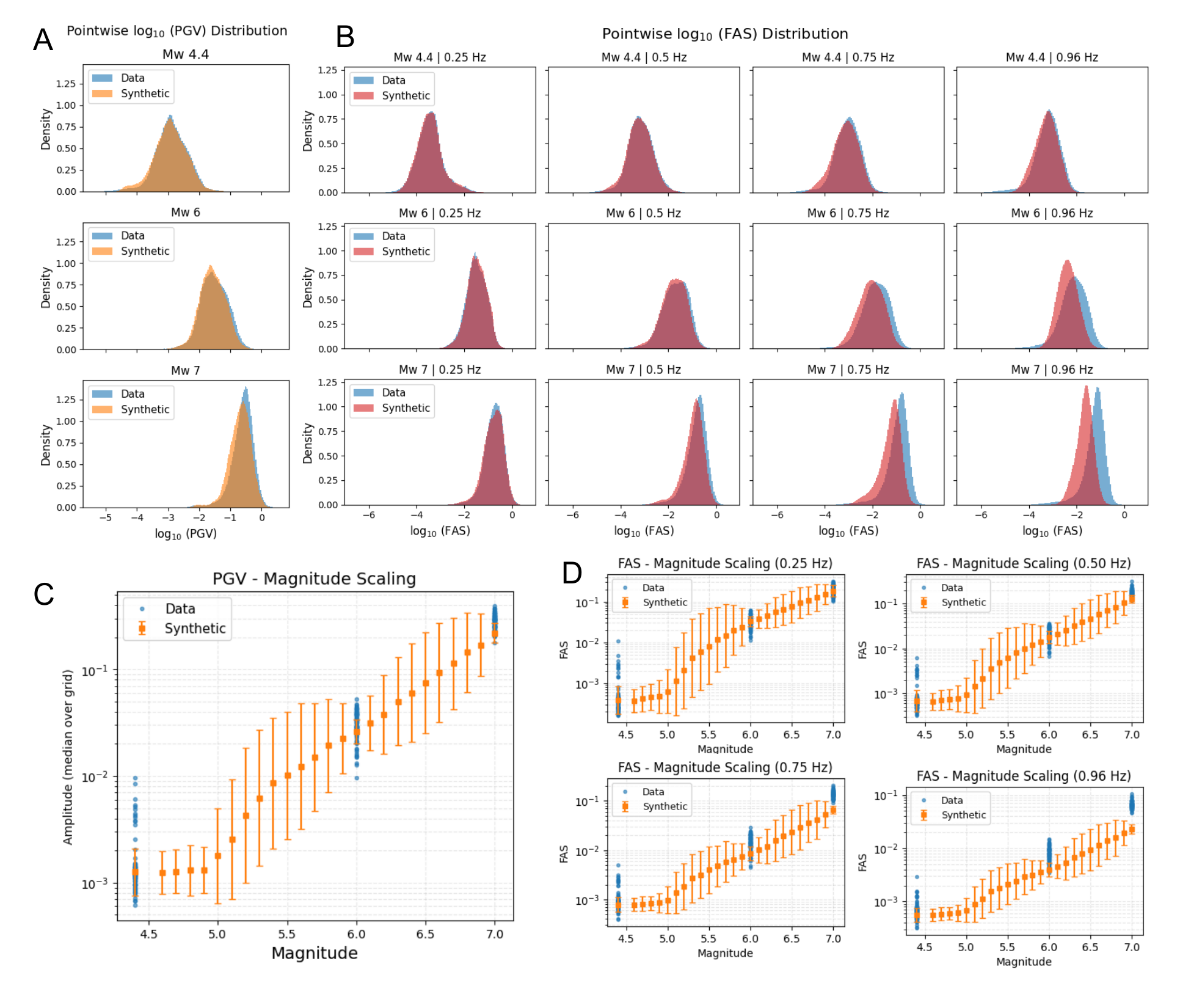}
    \caption{\textbf{Statistical distribution and magnitude scaling laws of generated ground motions}. \textbf{(A)} Pointwise distributions of $\log_{10}(\mathrm{PGV})$ for $M_w$ 4.4, 6.0, and 7.0 comparing held-out data to the same amount of synthetic realizations generated under identical conditions.
    \textbf{(B)} Pointwise distributions of $\log_{10}(\mathrm{FAS})$ at 0.25, 0.5, 0.75, and 0.96 Hz.
    \textbf{(C)} PGV magnitude scaling from $M_w$ 4.4--7.0 (0.1 magnitude increments), summarized by the spatial median per event over 100 realizations per bin.
    \textbf{(D)} Analogous magnitude scaling for FAS at the same frequencies; markers and error bars denote the ensemble geometric mean and standard deviation.}\label{fig:scaling}   
\end{figure*}

\begin{figure*}[ht]
    \vspace*{-.3cm}
    \centering

    \includegraphics[width=0.95\textwidth,trim={1cm, 0, 1cm, 0},clip]
    {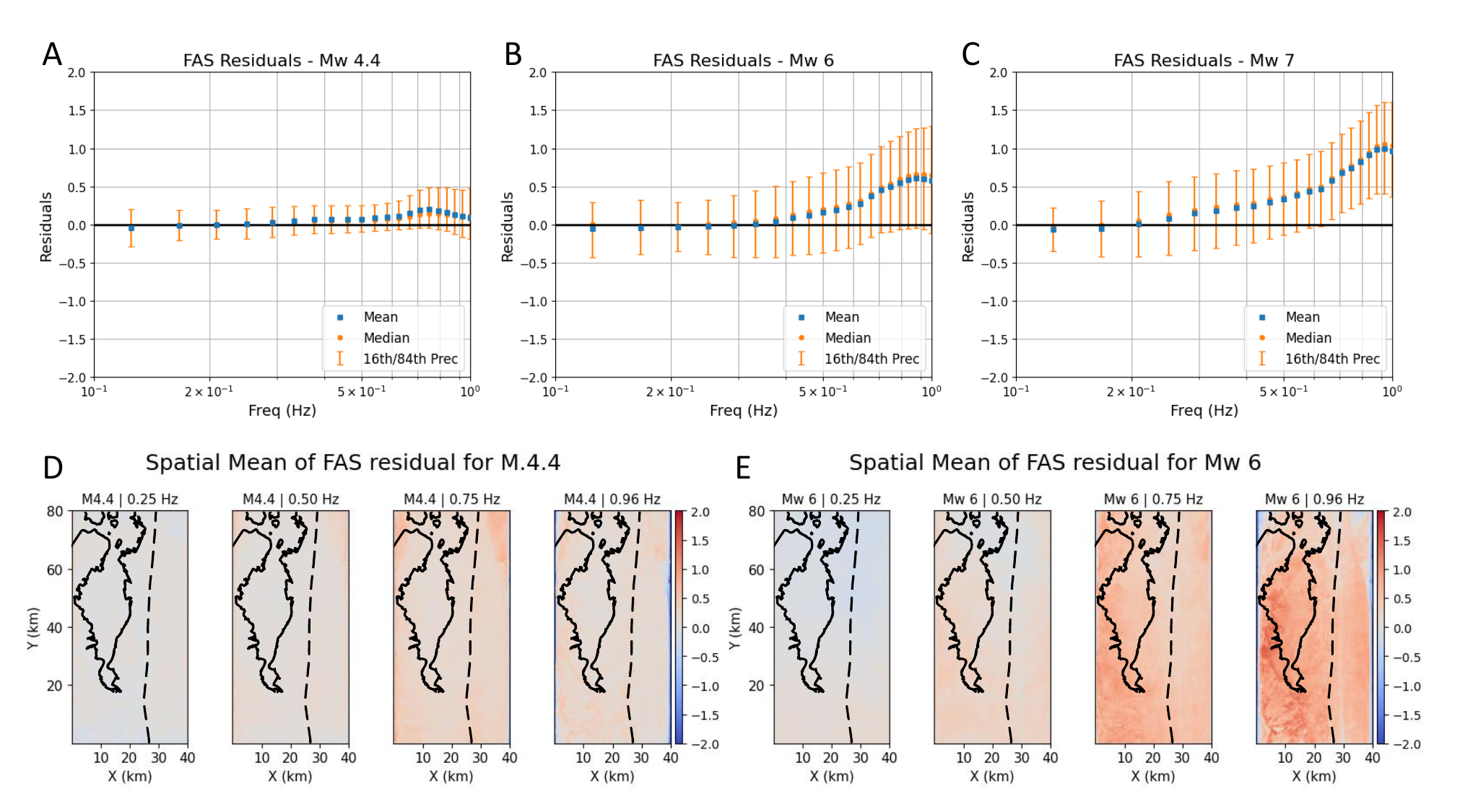}
\caption{\textbf{Frequency-dependent spectral residuals.}
\textbf{(A--C)} Residual statistics of horizontal Fourier amplitude spectra (FAS) as a function of frequency for $M_w$ 4.4, 6.0, and 7.0, computed over 300 test events using ensembles of 100 synthetic realizations per event; markers show the mean and median, and error bars indicate the 16th/84th percentiles.
\textbf{(D--E)} Spatial maps of the mean FAS residual at 0.25, 0.50, 0.75, and 0.96~Hz for $M_w$ 4.4 and $M_w$ 6.0 events. Black dash and solid lines represent Hayward fault and coastline, respectively.}
\label{fig:residual_analysis}   
\end{figure*}

\section{Results}



\subsection{Simulation Dataset and Evaluation Framework}
We evaluate the performance of \alg\ using a high-fidelity simulation dataset generated over the high-resolution domain of the U.S. Geological Survey San Francisco Bay Area three-dimensional seismic velocity model (SFVM). The computational domain spans $80 \times 40$\,km (Figure~\ref{fig:data_demo}).  The dataset comprises 5{,}300 simulated events, including 2{,}100 $M_w$~4.4 point-source earthquakes distributed across the entire domain, and 3{,}200 finite-rupture events with magnitudes $M_w$~6.0 and $M_w$~7.0 located on the Hayward Fault. 
The double-couple mechanism for the point-source events is based on the $M_w$~4.4 2018 Berkeley earthquake.
Model performance is assessed using a held-out test set of 300 events, stratified by magnitude class with 100 samples per class. All events were simulated using the SW4 platform~\cite{mccallen_eqsimmultidisciplinary_2021-1, mccallen_regional-scale_2024} using USGS v21.1 SFBA velocity model~\cite{aagaard_san_2021}, resolved to a maximum frequency of 1\,Hz. For each scenario, the surface velocity field was recorded over grid spacing, recorded on a $256 \times 128$ surface grid. Time histories were sampled at 4\,Hz over a duration of 24\,s, corresponding to 96 temporal snapshots. The resulting data tensor for each event has dimensions $[3, 256, 128, 96]$, representing the three velocity components over space and time.

We train \alg\ as a conditional generative model to learn the distribution of full spatiotemporal surface velocity wavefields conditioned on event parameters, specifically hypocenter location and moment magnitude. Model performance is evaluated using metrics that assess: (i) amplitude- and frequency-based statistics, including both ensemble mean behavior and predictive uncertainty; and (ii) the fidelity with which the model reproduces the underlying spatial correlation and spatiotemporal coherence structure. Additional details are provided in Sections~\ref{sec:data_prep} and~\ref{sec:metric}.

\subsection{Point-Source Scenario Evaluation}


We first evaluate \alg\ on a randomly selected representative $M_w~4.4$ point-source event from the held-out set. The model is conditioned on the corresponding hypocenter location and magnitude, and an ensemble of 100 stochastic realizations is generated to assess predictive performance and variability.

Figure~\ref{fig:point_source}A--B compares the amplitudes of the three-component velocity vector (Eq.~\ref{eqn:amp}) of the simulated event, treated as ground truth, with a randomly selected synthetic realization. Visually, \alg\ captures key features of the complex wavefield, including the primary wavefront geometry, localized scattering, and basin amplification (e.g., snapshot at 13\,s), all of which arise from wave propagation in a heterogeneous medium. The generated realization also preserves coherent spatiotemporal structure across the domain.

We quantify ensemble statistics using Peak Ground Velocity (PGV), Fourier Amplitude Spectra (FAS), and Normalized Cross-Correlation (NCC) (defined in Section~\ref{sec:metric}). 
PGV is used to evaluate the spatial distribution of ground-motion intensity (Figure~\ref{fig:point_source}C). The ensemble mean of the synthetic realizations closely matches the ground-truth field, indicating accurate recovery of large-scale amplitude patterns. The pixel-wise variability, quantified by the standard deviation across the 100 realizations, remains small for the point-source scenario, with a maximum value of $3.0 \times 10^{-4} \text{m/s}$, demonstrating low stochastic dispersion for this event class.
This occurs because all $M_w 4.4$ share the same point-source mechanism.
Figure~\ref{fig:point_source}D presents FAS maps at 0.25\,Hz and 0.75\,Hz to assess the model’s ability to reproduce frequency-dependent spatial patterns. Consistent with the PGV results, \alg\ accurately preserves the spatial distribution of spectral amplitude across frequencies, showing that \alg captures both low- and mid-frequency spatial structure within the resolved bandwidth.
Third, to assess spatiotemporal coherence and phase alignment, we compute NCC relative to a reference station. In Figure~\ref{fig:point_source}E, the reference station is placed at the geometric center of the domain.

From the NCC field, we extract two maps: the peak NCC coefficient (maximum similarity) and the corresponding time lag. Figure~\ref{fig:point_source}E shows close agreement between synthetic and ground-truth maps, indicating that the generated waveforms preserve both waveform similarity and phase alignment relative to the reference station. Importantly, the time-lag maps (bottom row) exhibit similar travel-time gradients, suggesting that \alg\ reproduces travel-time structure consistent with physically plausible wave propagation across the domain. Because a single central reference station may not fully characterize global coherence, we include additional NCC analyses with randomly selected reference stations in the Supplementary Materials, which show consistent behavior. Finally, we evaluate the wavefield propagation consistency using time--distance plots along two virtual station arrays (Profiles A1--A2 and B1--B2 in Figure~\ref{fig:point_source}F). As shown in Figure~\ref{fig:point_source}G--H, the synthetic patterns display similar arrival times and apparent velocities for P-waves, S-waves, and surface waves,  providing additional evidence of physically consistent regional wave propagation.

\subsection{Finite-Rupture Source Scenario Evaluation}

We next evaluate \alg\ on a representative $M_w\,7.0$ finite-rupture scenario on the Hayward Fault. Unlike the point-source case, these simulations involve finite-fault rupture with heterogeneous source-time-function realizations, posing a substantially more challenging generative task. Using the same evaluation protocol as in the point-source case, we first compare ground-truth wavefield snapshots with a representative synthetic realization (Figure~\ref{fig:extended_source}A--B). The model captures several key finite-source characteristics: the wavefield is no longer radially symmetric, but is instead shaped by rupture propagation along the fault trace (approximately in the $Y$-direction). In particular, \alg\ reproduces spatially asymmetric energy patterns consistent with rupture propagation and directivity, as well as basin-related scattering across the domain.
Figure~\ref{fig:extended_source}C shows that the synthetic mean PGV reproduces the high-intensity zone observed in the ground truth. 
The associated uncertainty map reveals elevated aleatory variability near the fault and in regions of strong geological contrast, consistent with learning physically meaningful variability rather than collapsing to a mean response. 
The accurate representation of the frequency content is supported by the FAS maps in Figure~\ref{fig:extended_source}D, which show that \alg preserves the spatial distribution of spectral energy and its decay behavior across the domain within the modeled frequency band. Spatiotemporal coherence is further assessed via NCC. The peak NCC coefficient maps (Figure~\ref{fig:extended_source}E) closely reproduce the main spatial decorrelation pattern in ground-truth: correlation decays with distance and medium heterogeneity, while remaining relatively high in the near field around the reference station. The corresponding time-lag maps exhibit similar travel-time gradients, suggesting that \alg synthesizes coherent wavefronts with realistic phase delays and apparent velocities under rupture-driven excitation.

Finally, time--distance plots along the virtual arrays (Figure~\ref{fig:extended_source}G--H) provide an additional evaluation of propagation behavior. Compared with the impulsive arrivals of the point-source case, the $M_w$\,7.0 wavefields exhibit prolonged wavetrains due to the finite rupture duration. \alg reproduces these extended phase arrivals and their moveout patterns, providing further evidence that the model captures the temporal evolution of large events. Collectively, these results show that \alg\ extends effectively to large-magnitude finite-rupture scenarios and captures key interactions between finite-fault source complexity and regional wave propagation.

\subsection{Site-Specific Spectral Evaluation with Predictive Uncertainty}


To assess local waveform fidelity beyond global spatial coherence, we examine site-specific FAS for three representative scenarios spanning the considered magnitude range (Figure~\ref{fig:site_spectra}A, C, E). For each scenario, we randomly select four stations covering a range of azimuths and source-to-site distances. For each event, we generate 100 realizations under identical conditioning and compute the ensemble geometric mean and standard deviation of both the horizontal power-mean FAS  (Eq.~\ref{eqn:fas}) and the vertical FAS.

As shown in Figure~\ref{fig:site_spectra}B, D, and F, the synthetic spectra (orange) closely match the ground truth (blue) over the analyzed 0.1--1\,Hz band for both horizontal and vertical components. \alg\ reproduces key spectral features, such as magnitude-dependent spectral shifts.The predicted geometric standard-deviation bands capture the overall spread of the ground-truth spectra across stations and magnitudes. Notably, the uncertainty for finite-rupture events ($M_w$\,6.0--7.0) is substantially larger than for the $M_w$\,4.4 point-source case, consistent with increased variability from source complexity, radiation pattern effects, and scattering in large events.

\subsection{Magnitude Scaling}\label{sec:mag_scale}

Beyond scenario-specific comparisons, we evaluate whether \alg\ reproduces the statistical distributions of ground motion and captures physically plausible scaling trends across magnitude. We first assess domain-wide ground-motion distribution by generating 300 synthetic realizations under the same conditioning as the held-out test events (for each event, we generate one synthetic). Figure~\ref{fig:scaling}A shows pointwise distributions of $\log_{10}(\mathrm{PGV})$ aggregated over the spatial domain for the three magnitude classes. The synthetic distributions (orange) closely match the ground truth (blue) distribution across all magnitudes. Figure~\ref{fig:scaling}B reports analogous comparisons for $\log_{10}(\mathrm{FAS})$ at four representative frequencies and shows broadly similar behavior. The 0.96\,Hz band is chosen to probe the high-frequency limit while avoiding artifacts near the strict 1\,Hz low-pass cutoff applied during training dataset preprocessing.

We next conduct a systematic scaling study (Figure~\ref{fig:scaling}C--D) by sampling magnitudes from $M_w$\,4.4 to 7.0 in increments of 0.1 and generating 100 realizations per magnitude bin. For each realization, we compute PGV and FAS ordinates at selected frequencies across the entire spatial domain, and then evaluate the corresponding mean and standard deviation.
The resulting plots report the ensemble mean and standard deviation of the intensity parameters for each magnitude level. The results exhibit a monotonic scaling trend consistent with physical expectations for increasing earthquake magnitude. Notably, the uncertainty (expressed as $\pm$ one standard deviation) is consistent with the training data for magnitudes ($M_w$~4.4, 6.0, and 7.0), while it broadens in the intermediate ranges where no direct training data are available. This behavior indicates reduced model confidence in interpolated (unseen) regimes while maintaining physically consistent scaling behavior.

A limitation appears in the high-frequency range for large-magnitude events. In the bottom-right panels of Figure~\ref{fig:scaling}D (0.75\,Hz and 0.96\,Hz), \alg\ underestimates the median amplitude for the $M_w$\,7.0 case relative to the ground truth, although the scaling trend is accurately reproduced up to $M_w$\,6.0. This discrepancy likely reflects the increased difficulty of representing high-frequency scattering associated with complex finite-fault ruptures within the current latent-space parameterization. 
One potential remedy would be to increase the dimensionality of the latent space in future training efforts; however, this would increase computational cost. Moreover, given the inherent uncertainty of numerical simulations at these higher frequencies, given the 3D velocity model resolution, such an increase was not pursued in the present study in order to maintain computational tractability. Instead, in future work, we aim to integrate synthetic simulations with recorded ground motions from real earthquakes, enabling the model to learn high-frequency characteristics directly from observational data.
Despite this limitation, the overall magnitude-scaling behavior remains physically consistent, with the ground-truth values generally falling within the predicted uncertainty bounds.

\subsection{Spectral Residual Analysis}

We perform a frequency-domain residual analysis (Eq.~\ref{eq:residual}) to quantify systematic bias. For each held-out test event, we generate an ensemble of 100 stochastic realizations under identical conditional parameters. Figure~\ref{fig:residual_analysis}A--C summarizes the aggregated residual statistics as a function of frequency (positive residuals indicate underestimation of spectral amplitude). For $M_w$\,4.4 events (Figure~\ref{fig:residual_analysis}A), residuals remain centered near zero across 0.1--1\,Hz, with a tight spread, indicating negligible bias. For finite-rupture events (Figure~\ref{fig:residual_analysis}B--C), the model maintains near-zero bias in the low-frequency range, but exhibits a systematic positive shift above $\sim$0.5\,Hz, reaching $\sim$0.5--1.0 (natural-log units) near 1\,Hz. This behavior is consistent with the previously noted limitation that \alg\ underestimates high-frequency spectral amplitudes for the large-magnitude scenarios. Given the deterministic nature of the velocity model and the fixed double-couple mechanism for the small-magnitude events, the predicted variability remains limited for these scenarios. In contrast, the broader uncertainty bands observed at larger magnitudes reflect the increased source complexity and enhanced wavefield scattering associated with finite-rupture events.

To assess the geographic distribution of these errors, we also examine spatial maps of the mean FAS residual over the domain. Figure~\ref{fig:residual_analysis}D--E shows residual fields for representative $M_w$\,4.4 and $M_w$\,6.0 cases. For the $M_w$\,4.4 case, the residual maps are nearly flat and centered around zero, indicating spatially unbiased predictions. For $M_w$\,6.0, the high-frequency residuals show the expected positive bias but remain relatively uniform, without pronounced spatial concentrations. These results suggest that the dominant error mode is a frequency-dependent amplitude bias rather than a strongly spatially localized modeling failure. 

To mitigate the observed frequency-dependent bias, we introduce an optional post-processing calibration step that applies a frequency-dependent amplitude correction to the generated spectra. This correction yields an approximately unbiased spectrum in expectation while preserving the predicted phase structure and spatiotemporal coherence.  As a result, agreement in FAS and magnitude-scaling behavior is improved. The full methodology and detailed evaluation of this calibration procedure are provided in the Supplementary Material. Importantly, this step should be viewed as a practical mitigation strategy rather than a substitute for improved generative modeling of high-frequency rupture processes and wavefield scattering.

\section{Discussion}
In this work, we introduced \alg, which to our knowledge is the first large-scale functional generative model for event-level regional ground-motion synthesis. \alg adopts a functional generative viewpoint, modeling the regional wavefield as a continuous spatiotemporal function. 

To achieve practical scalability, we combine neural operators with latent-space generation via flow matching and structure the model around a physics-inspired factorization through a physics-aligned subspace. The approach first generates a low-frequency wavefield that captures large-scale wave-propagation and spatial coherence effects, and then applies a super-resolution operator to reconstruct the missing high-frequency content. By restricting stochastic generation to this low-frequency, and decomposing synthesis into two stages—coherent low-frequency generation followed by super-resolution, \alg\ reduces both training and sampling complexity. This factorization further enables optional intermediate physical corrections at the low-frequency stage, ensuring that generated samples remain within a physically plausible solution manifold.

Evaluated against withheld physics-based simulations for the San Francisco Bay Area demonstrates that \alg\ captures key multiscale features relevant to hazard analysis, including coherent regional wavefront patterns, realistic spatial variability in intensity measures, and frequency-dependent scaling behavior consistent with the target distribution. Empirically, \alg\ delivers orders-of-magnitude speedups over conventional numerical simulation, opening a path toward uncertainty-aware regional hazard workflows for distributed infrastructure systems. Finally, we provide extensive ablation studies and sensitivity analyses in the Supplementary Material to examine the effects of key architectural and hyperparameter choices, with particular emphasis on the rationale for selecting a 0.75 Hz cutoff frequency.

\subsection{Limitations}

Despite the strong empirical performance, several limitations remain. First, while \alg\ performs well for point-source events across the modeled frequency band, large finite-rupture events exhibit systematic attenuation near the upper end of the band. As shown in Fig.~\ref{fig:residual_analysis}, the mean residual becomes increasingly positive above $\sim$0.5\,Hz, indicating that the current framework tends to underestimate high-frequency spectral amplitudes when rupture complexity and scattering are most pronounced. Extensive ablation studies in the Supplementary Material further show that this bias is not simply an artifact of the intermediate cutoff frequency. As a practical mitigation, we introduce an optional frequency-dependent calibration based on the mean spectral residual. Because the residual is approximately spatially uniform, the correction can be implemented as a single multiplicative scaling of each synthetic event’s spectral amplitudes across frequency, yielding an approximately unbiased spectrum in expectation. Although this adjustment improves engineering-relevant amplitude metrics, it does not replace the need for improved generative modeling of high-frequency rupture and scattering effects.

Second, although the functional formulation is in principle flexible, the current implementation relies on an FFT-based Fourier Neural Operator (FNO)~\cite{li_fourier_2021} trained on regular grids. This limits immediate applicability to complex geometries, topography-following or unstructured discretizations, and sparse irregular station networks, all of which are important for observation-driven hazard analysis. Addressing these limitations will require operator backbones that natively support irregular domains and point-set queries, together with training protocols that can fuse sparse observations and dense simulations within a unified functional generative framework.

Overall, we view \alg\ as a first step toward scalable generative modeling of regional ground-motion wavefields rather than as a deployment-ready end-to-end tool. Important challenges remain, including extension to broader frequency bands, richer finite-rupture physics, and principled calibration and adaptation to real recordings. Nevertheless, the framework demonstrates that uncertainty-aware regional synthesis can be made computationally tractable, opening a path toward next-generation hazard workflows for distributed infrastructure systems.

\subsection{Computational Efficiency and Comparative Analysis}

We evaluate the computational performance of \alg\ relative to both physics-based simulations and prior generative approaches. Given four scalar conditioning variables (three hypocenter coordinates and earthquake magnitude), \alg\ generates a stochastic realization of a three-component spatiotemporal wavefield on a default discretization of $[3,256,128,96]$. Each sample therefore contains approximately $9.4$ million scalar values in total, corresponding to about $3.1$ million space--time grid points with three waveform components at each point.

Once trained, \alg\ generates the full set of 5{,}300 events in 0.386 GPU-hours on a single workstation GPU (NVIDIA RTX 6000 Ada). In contrast, producing the same dataset with physics-based simulation required approximately 4{,}000 NVIDIA A100 GPU-hours and 6{,}200 AMD EPYC 7763 CPU-hours on a national supercomputing system (Section~\ref{sec:data_prep}), representing to a $>10^4\times$ reduction in compute. Because \alg\ models a continuous function, it also supports zero-shot super-resolution at inference time: without retraining, it can synthesize wavefields on a finer $[3,512,256,96]$ grid ($\sim 37.7$ million scalar values) in seconds on a single GPU; representative results are provided in the Supplementary Material.

Compared with prior generative models for single-site waveform synthesis, \alg\ addresses a substantially more challenging field-level generation problem. Methods such as cGM-GANO~\cite{shi_broadband_2024} and recent diffusion-based approaches~\cite{palgunadi_high_2025} generate three-component time histories at isolated sites but do not model within-event spatial correlations. Measured by total output dimensionality, the spatiotemporal wavefields generated by \alg\ are approximately $500\times$ larger than those of cGM-GANO and $780\times$ larger than the diffusion baseline, while being trained on a substantially smaller dataset than typically used for one-dimensional waveform models. In our framework, each earthquake scenario constitutes a single training sample representing a full regional wavefield, whereas in single-site waveform synthesis models each three-component ground motion at an individual station is treated as a separate sample.

A more closely related comparison is withWaveCastNet \cite{lyu_rapid_2025}, which also uses Bay Area physics-based simulations, but the problem setting is fundamentally different. WaveCastNet is a deterministic sequence to sequence forecasting model for earthquake early warning: it predicts future wavefields from an observed history window and introduces uncertainty only through expensive model ensembling. In contrast, \alg\ is a stochastic functional generative model for regional ground-motion synthesis conditioned only on low-dimensional physical parameters. Moreover, WaveCastNet is trained on small point-source earthquakes in a lower-frequency regime ($<0.5$\,Hz), whereas \alg\ trains a unified generator across both point-source and finite-rupture events up to 1\,Hz. Thus, WaveCastNet forecasts a realized event under strong observational constraints, whereas \alg\ must synthesize full event-level regional wavefields from physical conditions while matching spatial coherence, frequency content, and magnitude scaling.

\subsection{Difference from Video Generative Models}

Although both video generative models and \alg\ produce spatiotemporal outputs, the underlying problem setting is fundamentally different. Video models typically learn distributions over discrete frame tensors on fixed pixel grids and generate short clips conditioned on text, images, or preceding frames~\cite{ho_video_2022,he_latent_2023,wan_wan_2025}. In contrast, \alg\ models an event-level seismic wavefield as a random spatiotemporal function conditioned on physical parameters. This functional formulation supports resolution-flexible querying and is designed for scientific field synthesis rather than perceptual content generation.

The target distributions and constraints are also different. Video generation addresses broad, open-ended visual scenes, where realism is usually judged perceptually and strict physical consistency is not generally enforced. By contrast, \alg\ targets a specific physics-governed process (i.e., regional seismic wave propagation) that implicitly encodes a physical process. It incorporates this physical structure through a staged factorization and a Physics-Aligned Subspace that leads generation toward propagation-consistent solutions.

The data and evaluation regimes also differ accordingly. State-of-the-art video models are typically trained on internet-scale datasets with very large compute budgets, whereas \alg\ is designed for scientific-scale learning under limited data and compute while still generating outputs with millions of spatiotemporal degrees of freedom on a single workstation GPU. Furthermore, video generation is often assessed using perceptual realism and temporal consistency, while \alg\ is evaluated using physics- and engineering-driven criteria, including spectral fidelity, intensity measures, residual bias across frequency, and process-specific correlation structures.

\subsection{Future Directions: Irregular Geometry and Data Fusion}

A natural next step is to replace FFT-based FNO backbones with operator architectures that natively support irregular geometries, such as Mesh-Informed Neural Operators (MINO)~\cite{shi_mesh-informed_2025}. Unlike FFT-based operators, these architectures decouple evaluation locations from a fixed latent grid and can therefore support efficient inference on irregular station layouts, unstructured domains, and multi-resolution meshes. Extending \alg\ in this direction would improve compatibility with observational networks and move the framework closer to truly geometry-flexible functional generation.

A second priority is to develop hybrid training datasets and generative models that combine physics-based simulations with real observations. Neither source alone is sufficient for earthquake-engineering applications. Observational data provide the most reliable record of real ground motions, especially at high frequencies, but large-magnitude events relevant to infrastructure resilience are rare and station coverage is often too sparse to fully resolve the regional spatiotemporal structure of shaking. Physics-based simulations, by contrast, provide dense spatial coverage and controlled variation over earthquake scenarios, source properties, and receiver locations. However, they are often limited in frequency bandwidth because accurate propagation of high-frequency energy requires both substantial computational cost and highly detailed Earth models.

These complementary strengths suggest a hybrid generative framework that combines the spatial completeness and scenario control of simulations with the spectral realism of observations. The main challenge is that the two data sources differ not only in fidelity, but also in geometry, bandwidth, and error characteristics. One possible strategy is to use the Physics-Aligned Subspace as a shared low-frequency representation, where both simulations and observations can be compared or assimilated at scales that are more reliably resolved. Observations could then be used primarily to calibrate the high-frequency lifting (with super-resolution operator), improving the spectral realism of the generated fields while preserving the large-scale coherence learned from simulations. When combined with irregular-geometry operator backbones such as MINO, such a framework could support mixed-resolution and mixed-geometry training, and may provide a path toward regional generative models that retain simulation-scale coherence while better matching the high-frequency content of recorded ground motions. A systematic investigation of these directions is left to future work.

\subsection{Broader Impact: Toward Physics--AI Generative Hazard Modeling}

Beyond seismology, \alg\ points to a broader framework for generative modeling in natural-hazard science. Hurricanes, wildfires, tsunamis, and volcanic plumes differ in their governing physics, but they share a common modeling challenge: spatiotemporal dynamics on complex domains under uncertain forcing and incomplete observations~\cite{schwerdtner_uncertainty_2024,yu_probabilistic_2025,price_probabilistic_2025}. In this context, GMFlow shows how conditional solution operators can be learned directly in function space, complementing expensive grid-based simulation with fast, queryable synthesis of coherent spatiotemporal fields with uncertainty.

Accordingly, \alg\ can be viewed not only as a model for regional ground-motion synthesis, but also as a potential foundational framework for generative modeling of diverse large-scale spatiotemporal natural-hazard phenomena. By addressing the computational bottleneck of uncertainty-aware regional field generation, it could support large probabilistic scenario ensembles for downstream risk analysis of distributed infrastructure and emergency-planning workflows.

\section{Materials and Methods}

We first describe simulation dataset preparation, then formulate the problem of regional, event-level ground-motion synthesis in function space, followed by the design of the \alg\ architecture, training procedure, and inference workflow. Additional results, evaluation, and ablation are provided in the Supplementary Materials.

\subsection{Simulation Dataset Preparation}\label{sec:data_prep}

We selected the San Francisco Bay Area (SFBA) as our computational testbed and used the USGS v21.1 community velocity model to capture realistic heterogeneous geological structure~\cite{aagaard_san_2021}. This model has been rigorously evaluated using high-resolution 3D physics-based simulations and recorded seismicity. We generate simulated events with the SW4 finite-difference code, which solves the elastodynamic wave equations with fourth-order accuracy~\cite{mccallen_eqsimmultidisciplinary_2021-1, mccallen_regional-scale_2024}. The domain spans a $43 \times 92$ km region covering the Hayward Fault. Simulations use a minimum grid spacing of 31.25 m, a minimum shear-wave velocity of 250 m/s, absorbing boundaries, and are configured to resolve frequencies up to 1 Hz with at least eight points per wavelength throughout the mesh. Wavefields are stored on a sparse 250 m output grid. To simplify the learning setting and avoid boundary effects, we use a truncated $80 \times 40$ km subdomain with the same geometric center as the original computational domain, apply a sixth-order low-pass filter at 1 Hz, resample to $[256,128]$ spatial points, and truncate the duration to 24 seconds with a sampling frequency of 4 Hz, yielding three-component velocity wavefields of shape $[3,256,128,96]$.

The simulation dataset covers both small and large events with different source regimes. For small-magnitude earthquakes, we simulated 2{,}100 point-source events ($M_w$ 4.4) using a double-couple source with a Liu source-time function to model asymmetric energy ~\cite{liu_prediction_2006}. Hypocenter locations were sampled via Latin Hypercube Sampling to ensure adequate spatial coverage. For large-magnitude events, we targeted the Hayward Fault and generated 2{,}100 $M_w$ 6.0 scenarios and 1{,}100 $M_w$ 7.0 scenarios. These ruptures use the Graves--Pitarka kinematic model with stochastic slip distributions and rupture velocities to introduce realistic source heterogeneity. To improve computational efficiency for large-rupture events, we adopt a linear superposition strategy in which a finite-fault rupture is decomposed into a superposition of smaller rupture segments with appropriate time delays (Additional details are provided in Supplementary Materials Section B). The final dataset contains 5{,}000 training events (2{,}000 $M_w$ 4.4, 2{,}000 $M_w$ 6.0, and 1{,}000 $M_w$ 7.0) and 300 test events (100 per magnitude). Simulations were performed on Perlmutter at the National Energy Research Scientific Computing Center (NERSC) at Lawrence Berkeley National Laboratory, using a total of 4{,}000 NVIDIA A100 GPU-hours and 6{,}200 AMD EPYC 7763 (Milan) CPU-hours.

\subsection{Function-space formulation of conditional event-level ground-motion synthesis}

Let $D=\Omega\times[0,T]$ denote the spatiotemporal domain of wavefields where $\Omega\subset\mathbb{R}^2$ represents the spatial domain and $[0,T]$ the time domain.
We represent an an event-level three-component ground-motion velocity wavefield as a function $u: D \rightarrow \mathbb{R}^3$. We assume that $u$ belongs to a separable Hilbert space  $\mathcal{H}$. 
The probability distribution of wavefields is defined on the measurable space $(\mathcal{H},\mathcal{B}(\mathcal{H}))$, where $\mathcal{B}(\mathcal{H})$ denotes the Borel  $\sigma$-algebra on $\mathcal{H}$.
For an earthquake event described by conditioning variables $c$ (magnitude and source location), drawn from a distribution $c \sim \pi_c$, we assume that the wavefield follows an unknown conditional probability law $u \mid c \sim \mu_c^{\dagger}$. The marginal distribution of wavefields is therefore the mixture $\mu(A) = \int \mu_c^{\dagger}(A)\,\pi_c(dc)$ for any measurable set $A \in \mathcal{B}(\mathcal{H})$.
Our goal is to learn a conditional generative model $\mathcal{G}_\theta$ that approximates the family of conditional probability laws $\{\mu_c^{\dagger}\}$.

Let $z_0 \sim \pi_0$ denote a latent noise variable defined on the latent probability space 
$(\mathcal{Z},\mathcal{B}(\mathcal{Z}))$. Given conditioning variables $c$, we generate a synthetic
wavefield through the mapping $\hat u=\mathcal{G}_\theta(c,z_0)$, where $\mathcal{G}_\theta$ is a generator function, parameterized by $\theta$.
This construction induces, for each $c$, a probability measure $\mu_{\theta,c}$ on $(\mathcal{H},\mathcal{B}(\mathcal{H}))$ defined as the pushforward of the latent distribution
through the generator:
$
\mu_{\theta,c}:=(\mathcal{G}_\theta(c,\cdot))_{\#}\pi_0.
$
In other words, $\mu_{\theta,c}$ is the distribution of the random wavefield
$\hat{u} = \mathcal{G}_\theta(c,z_0)$ when $z_0 \sim \pi_0$.

We parameterize $\mathcal{G}_\theta$ using a neural operator and train it so that the model-induced conditional distributions $\mu_{\theta,c}$ approximate the true conditional
distributions $\mu_c^{\dagger}$ for conditioning variables $c$ drawn from the data distribution.

Adopting a function-space viewpoint for event-level ground-motion synthesis offers several advantages over standard fixed-grid generative modeling:

\begin{itemize}
    \item \textbf{Domain and Discretization Agnosticism.} The functional framework mitigates the "\textit{Curse of Dimensionality}" inherent to fixed-grid generative modeling. For regional wavefields observed on millions of spatiotemporal grid points, directly modeling a distribution on a single fixed discretization can be computationally and statistically burdensome. A function-space formulation provides a more flexible representation by separating the underlying continuous field from any particular discretization, where the discretization only introduces approximation errors. Further, this approach is also theoretically domain-agnostic, which supports domains with irregular boundaries that better aligns with the real-world earthquake scenario~\cite{shi_mesh-informed_2025}. 
    
    \item \textbf{Flexibility in Training and Inference.} In functional generative models, each training wavefield sample can be observed on its own unique set of spatial locations, ranging from dense regular simulation grids to sparsely sampled station networks from real earthquake, where the traditional fixed-grid generative theory falls short. During inference, the trained model can generate a continuous wavefield that can be evaluated at arbitrary query locations, enabling zero-shot generation on much finer discretizations not seen in the training dataset.
\end{itemize}

Despite these conceptual advantages, the function-space perspective alone does not resolve the practical computational challenges of large-scale generative modeling. Efficient implementation still requires careful design of the model architecture and training pipeline, which we describe in the following sections.

\subsection{Learning method and architecture}

\alg consists of three components organized in a two-stage synthesis pipeline. We elaborate each components in the following part. 

\label{sec:method}

\paragraph{Physics-Aligned Subspace and super-resolution operator.}
Let $A_f$ denote a temporal low-pass filtering operator with cutoff frequency $f_c$. We define the
\emph{Physics-Aligned Subspace} as
\[
\mathcal{H}_f := \operatorname{Range}(A_f) \subseteq \mathcal{H},
\qquad
u_f := A_f u \in \mathcal{H}_f.
\]

We refer to $\mathcal{H}_f$ as a \emph{Physics-Aligned Subspace} because it isolates the low-frequency component of the wavefield, which is typically the part most robustly determined by large-scale wave propagation physics and most reliably resolved in regional simulations. The higher-frequency content is more sensitive to the heterogeneity, scattering, source complexity, and discretization effects. Thus, the role of $A_f$ is not to impose a new physical constraint, but to define an intermediate representation aligned with the physically better-resolved scales of the problem, while deferring reconstruction of the remaining frequency content to a separate lifting stage. The super-resolution task is then to map the low-pass wavefield $u_f$ to a full-band reconstruction $\hat u \approx u$. We parameterize a super-resolution neural operator (SNO),
\[
S_\phi:\mathcal{H}_f \to \mathcal{H},
\qquad
\hat u := S_\phi(u_f),
\]
and optimize its parameters using paired samples $\{(u_f^{(i)},u^{(i)})\}_{i=1}^N$ by minimizing the $L^2$ reconstruction loss,
\[
\mathcal{L}_{\mathrm{SNO}}
=
\mathbb{E}_{u\sim\mu}\!\left[\|S_\phi(u_f)-u\|_{L^2(\mathcal{D})}^2\right]
\;\approx\;
\frac{1}{N}\sum_{i=1}^N \|S_\phi(u_f^{(i)})-u^{(i)}\|_{L^2(D)}^2,
\]
where $\|\cdot\|_{L^2(D)}$ is implemented by standard discrete quadrature on the chosen output discretization, i.e., mean-squared error over spatiotemporal grid points, and $S_\phi$ can be interpreted as the super-resolution operator, with the minimum mean-square-error estimator. Although $A_f$ is non-injective and therefore does not admit a unique inverse in general, the mapping $u_f \mapsto u$ can still be highly predictable on the support of physically realizable wavefields. In our fixed-region setting with a prescribed medium (the SFBA velocity model), the dominant variability is primarily source-driven: the low-frequency response retains large-scale source information, such as seismic moment and rupture directivity, which helps constrain the conditional statistics of the higher-frequency content.

In our experiments, we use $N=5{,}000$ paired training samples and set the cutoff frequency to $f_c=0.75$~Hz.
For computational efficiency, we evaluate $u_f$ on a coarser spatiotemporal discretization.
Let $(\mathbf{x, t})$ and $(\mathbf{x, t})_f$ denote the uniform grid coordinates associated with the fine and coarse discretizations of the same domain
$D$, with $ (\mathbf{x, t})\in\mathbb{R}^{M}$ and $(\mathbf{x, t})_f \in\mathbb{R}^{M_f}$, where
$M = 256\times128\times96$ and $M_f = 128\times64\times48$.
Accordingly, each sample is represented as
$u^{(i)}(\mathbf{x, t})\in\mathbb{R}^{3\times M}$ and $u_f^{(i)}((\mathbf{x, t})_f)\in\mathbb{R}^{3\times M_f}$. We parameterize $S_\phi$ with a residual Fourier Neural Operator (rFNO), which adds inter-block residual connections to a standard FNO backbone. To reduce dynamic-range issues across event magnitudes, we normalize both $u$ and $u_f$ by their sample-wise standard
deviation and append $\log_{10}$ of this scalar as an additional constant channel. The resulting inputs are
$u^{(i)}(x)\in\mathbb{R}^{4\times M}$ and $u_f^{(i)}(x_f)\in\mathbb{R}^{4\times M_f}$.
This normalization is important because waveform amplitudes can vary substantially with magnitude; without rescaling,
small events may be taken as noise or negelected during training relative to large events, degrading performance across the magnitude range.

\paragraph{Autoencoding Neural Operator.}
Inspired by latent diffusion frameworks~\cite{rombach_high-resolution_2022} and U-shape neural operator (UNO)~\cite{rahman_u-no_2023}, we introduce an Autoencoding Neural Operator (AENO) to learn a compact, fixed-size latent representation of the low-frequency wavefield $u_f$. Conceptually, AENO maps the low-frequency wavefield to a low-dimensional latent tensor, reducing the computational cost of the downstream generative model by allowing flow matching to operate in a compressed latent space.

Formally, the AENO consists of an encoder operator $\mathcal{E}_{\psi}$ and a decoder operator $\mathcal{D}_{\psi}$, parameterized by $\psi$. Given an input $u_f \sim \mu_f$, where $\mu_f = (A_f)_{\#}\mu$ is the pushforward of the original measure $\mu$ under the low-pass filter $A_f$, the encoder maps $u_f$ to a latent variable 
$
z_1 = \mathcal{E}_{\psi}(u_f) \in \mathbb{R}^{1 \times 32 \times 16 \times 16}.
$
Thus, the latent representation has a single channel and fixed spatiotemporal dimensions. In principle, because $\mathcal{E}_{\psi}$ is formulated as an operator, it can be applied across varying input discretizations while mapping them into the same latent space $\mathcal{Z}$. The decoder then reconstructs the low-frequency wavefield from the latent representation
$
\hat{u}_f = \mathcal{D}_{\psi}(z_1).
$

We optimize $\psi$ using an $L^2$ reconstruction objective over the data distribution:
\[
\mathcal{L}_{\mathrm{AENO}}
=
\mathbb{E}_{u_f \sim \mu_f}\!\left[
\big\|\mathcal{D}_{\psi}(\mathcal{E}_{\psi}(u_f)) - u_f\big\|_{L^2(D)}^2
\right]
\;\approx\;
\frac{1}{N}\sum_{i=1}^N
\big\|\mathcal{D}_{\psi}(\mathcal{E}_{\psi}(u_f^{(i)})) - u_f^{(i)}\big\|_{L^2(D)}^2.
\]
As in the SNO objective, the $L^2(D)$ norm is implemented discretely on the chosen training grid. Although the AENO framework can in principle support variable input discretizations, in our experiments we evaluate $u_f$ on a fixed coarse grid for computational efficiency. We instantiate both $\mathcal{E}{\psi}$ and $\mathcal{D}{\psi}$ with the UNO architecture using rFNO blocks, for consistency with the SNO implementation, and additionally apply Instance Normalization~\cite{ulyanov_instance_2017} after each block to stabilize latent activations.S

\paragraph{Conditional flow matching with clean latent prediction.}
Let $z_1 \sim \pi_1$ denote a latent data sample obtained by encoding a physics-aligned wavefield $u_f$ via
$
z_1 = \mathcal{E}_{\psi}(u_f),
$
where $z_1 \in \mathbb{R}^{32\times 16\times 16}$ in our implementation. We sample the base noise as
$
z_0 \sim \pi_0 := \mathcal{N}(0,I).
$
Rectified flow~\cite{liu_flow_2022} (equivalently, independent-coupling flow matching~\cite{tong_improving_2024}) defines a linear coupling path between $z_0$ and $z_1$,
$
z_t := (1-t)\,z_0 + t\,z_1$, where $ t\in[0,1],
$
with constant marginal target velocity along the path,
\[
v_t(z_t) := \frac{d}{dt}z_t = z_1 - z_0.
\]

We parameterize a conditional velocity field $v_{\theta}$ and train it with the conditional rectified-flow objective
\[
\mathcal{L}_{\text{FM}} = \mathbb{E}_{c\sim\pi_c,\ z_1\sim\pi_1(\cdot\mid c),\ z_0\sim\pi_0,\ t\sim\mathcal{U}[0,1]}
\Big[\big\|v_{\theta}(z_t,t,c) - (z_1-z_0)\big\|_2^2\Big].
\]

Following recent practice, instead of predicting $v_{\theta}$ directly, we adopt \emph{clean-prediction with velocity-loss}: under a manifold assumption, clean samples lie on a low-dimensional data manifold where directly predicting the clean samples and then recalculate the velocity yields better performance~\cite{li_back_2026}. Specifically, we parameterize a neural operator $\mathcal{F}_{\theta}$ and set
\[
\hat z_1 = \mathcal{F}_{\theta}(z_t,t,c),
\qquad
v_{\theta}(z_t,t,c) = \frac{\hat z_1 - z_t}{1-t},
\]
for $t\in[0,1]$ (denominator is clipped around t=1). At inference time, given conditioning variables $c$, we sample $z_0\sim\pi_0$ and integrate the conditional ODE
\[
\frac{d z_t}{dt} = v_{\theta}(z_t,t,c),
\qquad z_{t=0}=z_0,
\]
to obtain $z_1 = z_{t=1}$. In practice, we solve this ODE numerically and use a 50-step Euler solver by default.

\paragraph{End-to-end inference.}\label{sec:inference}
Given conditioning variables $c$, we first sample a latent realization $\hat z_1$ by integrating the conditional rectified-flow ODE. We then decode $\hat z_1$ to obtain a low-frequency wavefield
$
\hat u_f = \mathcal{D}_{\psi}(\hat z_1).
$
Because the Physics-Aligned Subspace is defined by an explicit cutoff frequency ($f_c=0.75$\,Hz), we can optionally enforce this band limit at inference time to suppress spurious energy above $f_c$ in the intermediate stage, for example by applying an additional low-pass projection. In our experiments, this extra step is not required, as the generated $\hat u_f$ already satisfies the prescribed constraint to a good level. Finally, we recover a broader-band wavefield, resolved up to 1\,Hz and sampled at 4\,Hz, using the super-resolution neural operator,
$
\hat u := S_{\phi}(\hat u_f).
$
Overall, the inference pipeline is
\[
z_0 \sim \pi_0
\;\xrightarrow[\text{rectified flow}]{\,v_{\theta}(\cdot,\cdot,c)\,}
\hat z_1
\;\xrightarrow[\text{decode}]{\,\mathcal{D}_{\psi}\,}
\hat u_f
\;\xrightarrow[\text{super-resolve}]{\,S_{\phi}\,}
\hat u.
\]

\subsection{Metrics for Evaluation}\label{sec:metric}

We evaluate \alg using standard engineering metrics that quantify both pointwise intensity and spatiotemporal structure. For a three-component wavefield $u(x,t)$, we denote the two horizontal components by $u_{h_1},u_{h_2}$ and the vertical component by $u_v$. The instantaneous 3C amplitude is
\begin{equation}
\|u(x,t)\|_2=\sqrt{u_{h_1}(x,t)^2+u_{h_2}(x,t)^2+u_v(x,t)^2}, \qquad \forall (x,t)\in D.
\label{eqn:amp}
\end{equation}

\paragraph{Peak ground velocity (PGV).}
We define PGV at location $x$ as the maximum 3C amplitude over time,
\begin{equation}
\mathrm{PGV}(x)=\max_{t\in[0,T]}\|u(x,t)\|_2, \qquad \forall x\in\Omega.
\end{equation}

\paragraph{Fourier amplitude spectrum (FAS).}
Let $A_{h_1}(x,f)$ and $A_{h_2}(x,f)$ denote the Fourier amplitude spectra of the horizontal traces
$u_{h_1}(x,[0,T])$ and $u_{h_2}(x,[0,T])$ at frequency $f$. We report the power-mean horizontal FAS,
\begin{equation}
A_h(x,f)=\sqrt{\frac{1}{2}\left(A_{h_1}(x,f)^2+A_{h_2}(x,f)^2\right)}.
\label{eqn:fas}
\end{equation}

\paragraph{Peak normalized cross-correlation (NCC).}
Let $x_0$ be a reference location with reference traces
$r_{h_1}(t)=u_{h_1}(x_0,t)$, $r_{h_2}(t)=u_{h_2}(x_0,t)$, and $r_v(t)=u_v(x_0,t)$.
For each location $x$ and lag $k\in\{-K,\ldots,K\}$, define the valid overlap index set
\[
\mathcal{T}_k=
\begin{cases}
\{0,\ldots,T-1-k\}, & k\ge 0,\\
\{-k,\ldots,T-1\}, & k<0,
\end{cases}
\]
and the 3C normalized cross-correlation
\begin{equation}
\rho(x,k)=
\frac{
\sum\limits_{t\in\mathcal{T}_k}\!\left[
u_{h_1}(x,t+k)\,r_{h_1}(t)
+u_{h_2}(x,t+k)\,r_{h_2}(t)
+u_v(x,t+k)\,r_v(t)
\right]
}{
\sqrt{
\left(\sum\limits_{t\in\mathcal{T}_k}[r_{h_1}^2(t)+r_{h_2}^2(t)+r_v^2(t)]\right)
\left(\sum\limits_{t\in\mathcal{T}_k}[u_{h_1}^2(x,t+k)+u_{h_2}^2(x,t+k)+u_v^2(x,t+k)]\right)
}
}.
\end{equation}
The peak NCC coefficient $\rho^\ast(x)$ and corresponding lag $\tau^\ast(x)$ are
\begin{equation}
\rho^\ast(x)=\max_{k\in[-K,K]}\rho(x,k), \qquad
k^\ast(x)=\arg\max_{k\in[-K,K]}\rho(x,k), \qquad
\tau^\ast(x)=k^\ast(x)\,\Delta t,
\end{equation}
where $\Delta t=0.25~\mathrm{s}$ is the sampling interval and
\[
K=\min\!\left(\left\lfloor \frac{\mathrm{max\_lag\_s}}{\Delta t}+\tfrac12\right\rfloor,\;T-1\right).
\]
In this work, we choose $\mathrm{max\_lag\_s}=6~\mathrm{s}$.

\paragraph{Interpolated conditions for magnitude scaling.}\label{sec:scaling_gen}
Since \alg requires conditions as input for generating synthetic events, thus for the magnitude scaling section ~\ref{sec:mag_scale}. Interpolation conditions are built for target magnitudes from 4.6 to 6.9 (step 0.1), with 100 samples per target magnitude.  
For each target magnitude \(m\), the code linearly mixes two neighboring source pools: m4.4--m6 when \(m \le 6.0\), and m6--m7 when \(m > 6.0\).  
The mixing weight is computed as
\[
\alpha=\frac{m-4.4}{6.0-4.4}\quad (m\le 6.0), 
\qquad
\alpha=\frac{m-6.0}{7.0-6.0}\quad (m>6.0).
\]
Then it randomly samples the required number from each pool (without replacement), copies the hypocenter from those condition vectors.

\paragraph{Spectral residuals.}
To quantify frequency-dependent bias, for each test event we generate 100 synthetic realizations under the same conditioning variables (e.g., magnitude and hypocenter). Let $A_h^{\mathrm{data}}(x,f)$ and $A_h^{\mathrm{syn}}(x,f)$ denote the horizontal FAS of the observed and synthetic wavefields, respectively. We compute the ensemble mean in log space and define the residual
\begin{equation}
\epsilon(x, f) = \ln(A_h^{\text{data}}(x, f)) - \text{mean}(\ln(A_h^{\text{syn}}(x, f))
\label{eq:residual}
\end{equation}

\section{Data and code availability}
To promote transparency, reproducibility, and broader community use, all data, code, and pretrained models associated with this study are publicly available. The raw earthquake simulation dataset for the San Francisco Bay Area (approximately 3.3~TB) is hosted at \url{https://huggingface.co/datasets/Yaozhong/GMFlow_raw_simulation} and represents, to our knowledge, one of the largest open-source earthquake ground-motion simulation datasets currently available. The processed dataset and pretrained model weights are available at \url{https://huggingface.co/datasets/Yaozhong/GMFlow}, and the source code is available at \url{https://github.com/yzshi5/GMFlow}. Together, these resources not only enable full reproduction of the results reported in this study, but also support future research in earthquake engineering, scientific machine learning, and broader generative modeling applications in science and engineering.

\section{Supplementary Materials}

\textbf{This PDF file includes:}\\
Supplementary Text \\
Figs. S1 to S13 \\
Tables. T1 to T4 \\
Legends for movies. M1 to M6

\textbf{Other Supplementary Material for this manuscript includes the following:}\\
Movies. M1 to M6


\bibliographystyle{unsrt}  
\bibliography{GMFlow.bib}

@article{palgunadi_high_2025,
	title = {High {Resolution} {Seismic} {Waveform} {Generation} {Using} {Denoising} {Diffusion}},
	volume = {2},
	issn = {2993-5210},
	url = {https://onlinelibrary.wiley.com/doi/abs/10.1029/2025JH000862},
	doi = {10.1029/2025JH000862},
	language = {en},
	number = {4},
	urldate = {2025-12-31},
	journal = {Journal of Geophysical Research: Machine Learning and Computation},
	author = {Palgunadi, Kadek Hendrawan and Bergmeister, Andreas and Bosisio, Andrea and Ermert, Laura A. and Koroni, Maria and Perraudin, Nathanaël and Dirmeier, Simon and Meier, Men-Andrin},
	year = {2025},
	note = {\_eprint: https://agupubs.onlinelibrary.wiley.com/doi/pdf/10.1029/2025JH000862},
	pages = {e2025JH000862},
}

@misc{shi_stochastic_2025,
	title = {Stochastic {Process} {Learning} via {Operator} {Flow} {Matching}},
	url = {http://arxiv.org/abs/2501.04126},
	doi = {10.48550/arXiv.2501.04126},
	urldate = {2025-12-31},
	publisher = {arXiv},
	author = {Shi, Yaozhong and Ross, Zachary E. and Asimaki, Domniki and Azizzadenesheli, Kamyar},
	month = oct,
	year = {2025},
	note = {arXiv:2501.04126 [cs]},
}

@misc{shi_mesh-informed_2025,
	title = {Mesh-{Informed} {Neural} {Operator} : {A} {Transformer} {Generative} {Approach}},
	shorttitle = {Mesh-{Informed} {Neural} {Operator}},
	url = {http://arxiv.org/abs/2506.16656},
	doi = {10.48550/arXiv.2506.16656},
	urldate = {2025-12-31},
	publisher = {arXiv},
	author = {Shi, Yaozhong and Ross, Zachary E. and Asimaki, Domniki and Azizzadenesheli, Kamyar},
	month = oct,
	year = {2025},
	note = {arXiv:2506.16656 [cs]},
}

@article{guidotti_modeling_2016,
	title = {Modeling the resilience of critical infrastructure: {The} role of network dependencies},
	volume = {1},
	issn = {2378-9689},
	number = {3-4},
	journal = {Sustainable and resilient infrastructure},
	publisher = {Taylor \& Francis},
	author = {Guidotti, Roberto and Chmielewski, Hana and Unnikrishnan, Vipin and Gardoni, Paolo and McAllister, Therese and van de Lindt, John},
	year = {2016},
	pages = {153--168},
}

@article{shi_broadband_2024,
	title = {Broadband ground‐motion synthesis via generative adversarial neural operators: {Development} and validation},
	volume = {114},
	issn = {0037-1106},
	number = {4},
	journal = {Bulletin of the Seismological Society of America},
	publisher = {Seismological Society of America},
	author = {Shi, Yaozhong and Lavrentiadis, Grigorios and Asimaki, Domniki and Ross, Zachary E and Azizzadenesheli, Kamyar},
	year = {2024},
	pages = {2151--2171},
}

@article{pitarka_broadband_2013,
	title = {Broadband {Ground}‐{Motion} {Simulation} of an {Intraslab} {Earthquake} and {Nonlinear} {Site} {Response}: 2010 {Ferndale}, {California}, {Earthquake} {Case} {Study}},
	volume = {84},
	issn = {0895-0695},
	shorttitle = {Broadband {Ground}‐{Motion} {Simulation} of an {Intraslab} {Earthquake} and {Nonlinear} {Site} {Response}},
	url = {https://doi.org/10.1785/0220130031},
	doi = {10.1785/0220130031},
	number = {5},
	urldate = {2026-01-03},
	journal = {Seismological Research Letters},
	author = {Pitarka, Arben and Thio, Hong Kie and Somerville, Paul and Bonilla, Luis Fabian},
	month = sep,
	year = {2013},
	pages = {785--795},
}

@article{rodgers_regionalscale_2020,
	title = {Regional‐{Scale} {3D} {Ground}‐{Motion} {Simulations} of {Mw} 7 {Earthquakes} on the {Hayward} {Fault}, {Northern} {California} {Resolving} {Frequencies} 0–10 {Hz} and {Including} {Site}‐{Response} {Corrections}},
	volume = {110},
	issn = {0037-1106},
	url = {https://doi.org/10.1785/0120200147},
	doi = {10.1785/0120200147},
	number = {6},
	urldate = {2026-01-03},
	journal = {Bulletin of the Seismological Society of America},
	author = {Rodgers, Arthur J. and Pitarka, Arben and Pankajakshan, Ramesh and Sjögreen, Bjorn and Petersson, N. Anders},
	month = aug,
	year = {2020},
	pages = {2862--2881},
}

@article{azizzadenesheli_neural_2024,
	title = {Neural operators for accelerating scientific simulations and design},
	volume = {6},
	copyright = {2024 Springer Nature Limited},
	issn = {2522-5820},
	url = {https://www.nature.com/articles/s42254-024-00712-5},
	doi = {10.1038/s42254-024-00712-5},
	language = {en},
	number = {5},
	urldate = {2026-01-04},
	journal = {Nature Reviews Physics},
	publisher = {Nature Publishing Group},
	author = {Azizzadenesheli, Kamyar and Kovachki, Nikola and Li, Zongyi and Liu-Schiaffini, Miguel and Kossaifi, Jean and Anandkumar, Anima},
	month = may,
	year = {2024},
	pages = {320--328},
}

@article{florez_datadriven_2022,
	title = {Data‐{Driven} {Synthesis} of {Broadband} {Earthquake} {Ground} {Motions} {Using} {Artificial} {Intelligence}},
	volume = {112},
	issn = {0037-1106},
	url = {https://doi.org/10.1785/0120210264},
	doi = {10.1785/0120210264},
	number = {4},
	urldate = {2026-01-04},
	journal = {Bulletin of the Seismological Society of America},
	author = {Florez, Manuel A. and Caporale, Michaelangelo and Buabthong, Pakpoom and Ross, Zachary E. and Asimaki, Domniki and Meier, Men‐Andrin},
	month = apr,
	year = {2022},
	pages = {1979--1996},
}

@article{aquib_broadband_2024,
	title = {Broadband {Ground}‐{Motion} {Simulations} with {Machine}‐{Learning}‐{Based} {High}‐{Frequency} {Waves} from {Fourier} {Neural} {Operators}},
	volume = {114},
	issn = {0037-1106},
	url = {https://doi.org/10.1785/0120240027},
	doi = {10.1785/0120240027},
	number = {6},
	urldate = {2026-01-04},
	journal = {Bulletin of the Seismological Society of America},
	author = {Aquib, Tariq Anwar and Mai, P. Martin},
	month = sep,
	year = {2024},
	pages = {2846--2868},
}

@article{lyu_rapid_2025,
	title = {Rapid wavefield forecasting for earthquake early warning via deep sequence to sequence learning},
	volume = {16},
	copyright = {2025 The Author(s)},
	issn = {2041-1723},
	url = {https://www.nature.com/articles/s41467-025-65435-2},
	doi = {10.1038/s41467-025-65435-2},
	language = {en},
	number = {1},
	urldate = {2026-01-06},
	journal = {Nature Communications},
	publisher = {Nature Publishing Group},
	author = {Lyu, Dongwei and Nakata, Rie and Ren, Pu and Mahoney, Michael W. and Pitarka, Arben and Nakata, Nori and Erichson, N. Benjamin},
	month = nov,
	year = {2025},
	pages = {10622},
}

@misc{li_fourier_2021,
	title = {Fourier {Neural} {Operator} for {Parametric} {Partial} {Differential} {Equations}},
	url = {http://arxiv.org/abs/2010.08895},
	doi = {10.48550/arXiv.2010.08895},
	urldate = {2026-02-06},
	publisher = {arXiv},
	author = {Li, Zongyi and Kovachki, Nikola and Azizzadenesheli, Kamyar and Liu, Burigede and Bhattacharya, Kaushik and Stuart, Andrew and Anandkumar, Anima},
	month = may,
	year = {2021},
	note = {arXiv:2010.08895 [cs]},
}

@article{pinilla-ramos_performance_2025,
	title = {Performance evaluation of the {USGS} velocity model for the {San} {Francisco} {Bay} {Area}},
	volume = {41},
	issn = {8755-2930},
	number = {1},
	journal = {Earthquake Spectra},
	publisher = {SAGE Publications Sage UK: London, England},
	author = {Pinilla-Ramos, Camilo and Pitarka, Arben and McCallen M. EERI, David and Nakata, Rie},
	year = {2025},
	pages = {457--494},
}

@article{aagaard_san_2021,
	title = {San {Francisco} {Bay} region {3D} seismic velocity model v21. 1},
	journal = {US Geological Survey (USGS) Data Release},
	author = {Aagaard, Brad T and Hirakawa, Evan T},
	year = {2021},
	pages = {1310},
}

@article{graves_kinematic_2016,
	title = {Kinematic {Ground}‐{Motion} {Simulations} on {Rough} {Faults} {Including} {Effects} of {3D} {Stochastic} {Velocity} {Perturbations}},
	volume = {106},
	issn = {0037-1106},
	url = {https://doi.org/10.1785/0120160088},
	doi = {10.1785/0120160088},
	number = {5},
	urldate = {2026-02-17},
	journal = {Bulletin of the Seismological Society of America},
	author = {Graves, Robert and Pitarka, Arben},
	month = aug,
	year = {2016},
	pages = {2136--2153},
}

@article{liu_prediction_2006,
	title = {Prediction of broadband ground-motion time histories: {Hybrid} low/high-frequency method with correlated random source parameters},
	volume = {96},
	issn = {1943-3573},
	number = {6},
	journal = {Bulletin of the Seismological Society of America},
	publisher = {Seismological Society of America},
	author = {Liu, Pengcheng and Archuleta, Ralph J and Hartzell, Stephen H},
	year = {2006},
	pages = {2118--2130},
}

@article{mccallen_eqsimmultidisciplinary_2021,
	title = {{EQSIM}—{A} multidisciplinary framework for fault-to-structure earthquake simulations on exascale computers part {I}: {Computational} models and workflow},
	volume = {37},
	issn = {8755-2930},
	number = {2},
	journal = {Earthquake Spectra},
	publisher = {SAGE Publications Sage UK: London, England},
	author = {McCallen, David and Petersson, Anders and Rodgers, Arthur and Pitarka, Arben and Miah, Mamun and Petrone, Floriana and Sjogreen, Bjorn and Abrahamson, Norman and Tang, Houjun},
	year = {2021},
	pages = {707--735},
}

@article{mccallen_eqsimmultidisciplinary_2021-1,
	title = {{EQSIM}—{A} multidisciplinary framework for fault-to-structure earthquake simulations on exascale computers, part {II}: {Regional} simulations of building response},
	volume = {37},
	issn = {8755-2930},
	number = {2},
	journal = {Earthquake Spectra},
	publisher = {SAGE Publications Sage UK: London, England},
	author = {McCallen, David and Petrone, Floriana and Miah, Mamun and Pitarka, Arben and Rodgers, Arthur and Abrahamson, Norman},
	year = {2021},
	pages = {736--761},
}

@article{mccallen_regional-scale_2024,
	title = {Regional-scale fault-to-structure earthquake simulations with the {EQSIM} framework: {Workflow} maturation and computational performance on {GPU}-accelerated exascale platforms},
	volume = {40},
	issn = {8755-2930},
	number = {3},
	journal = {Earthquake Spectra},
	publisher = {SAGE Publications Sage UK: London, England},
	author = {McCallen, David and Pitarka, Arben and Tang, Houjun and Pankajakshan, Ramesh and Petersson, N Anders and Miah, Mamun and Huang, Junfei},
	year = {2024},
	pages = {1615--1652},
}

@article{mccallen_open-access_2025,
	title = {An open-access simulated earthquake ground-motion database for an {M7} {Hayward} {Fault} earthquake in the {San} {Francisco} {Bay} {Region}},
	volume = {41},
	issn = {8755-2930},
	number = {3},
	journal = {Earthquake Spectra},
	publisher = {SAGE Publications Sage UK: London, England},
	author = {McCallen, David and Pitarka, Arben and Tang, Houjun and Nakata, Rie and Mosalam, Khalid M and Petrone, Floriana and Günay, Selim and Perez, Claudio},
	year = {2025},
	pages = {2560--2597},
}

@article{pitarka_refinements_2022,
	title = {Refinements to the {Graves}–{Pitarka} kinematic rupture generator, including a dynamically consistent slip‐rate function, applied to the 2019 {M} w 7.1 {Ridgecrest} earthquake},
	volume = {112},
	issn = {0037-1106},
	number = {1},
	journal = {Bulletin of the Seismological Society of America},
	publisher = {Seismological Society of America},
	author = {Pitarka, Arben and Graves, Robert and Irikura, Kojiro and Miyakoshi, Ken and Wu, Changjiang and Kawase, Hiroshi and Rodgers, Arthur and McCallen, David},
	year = {2022},
	pages = {287--306},
}

@misc{rahman_u-no_2023,
	title = {U-{NO}: {U}-shaped {Neural} {Operators}},
	shorttitle = {U-{NO}},
	url = {http://arxiv.org/abs/2204.11127},
	doi = {10.48550/arXiv.2204.11127},
	urldate = {2026-02-17},
	publisher = {arXiv},
	author = {Rahman, Md Ashiqur and Ross, Zachary E. and Azizzadenesheli, Kamyar},
	month = may,
	year = {2023},
	note = {arXiv:2204.11127 [cs]},
}

@misc{liu_flow_2022,
	title = {Flow {Straight} and {Fast}: {Learning} to {Generate} and {Transfer} {Data} with {Rectified} {Flow}},
	shorttitle = {Flow {Straight} and {Fast}},
	url = {http://arxiv.org/abs/2209.03003},
	doi = {10.48550/arXiv.2209.03003},
	urldate = {2026-02-17},
	publisher = {arXiv},
	author = {Liu, Xingchao and Gong, Chengyue and Liu, Qiang},
	month = sep,
	year = {2022},
	note = {arXiv:2209.03003 [cs]},
}

@misc{tong_improving_2024,
	title = {Improving and generalizing flow-based generative models with minibatch optimal transport},
	url = {http://arxiv.org/abs/2302.00482},
	doi = {10.48550/arXiv.2302.00482},
	urldate = {2026-02-17},
	publisher = {arXiv},
	author = {Tong, Alexander and Fatras, Kilian and Malkin, Nikolay and Huguet, Guillaume and Zhang, Yanlei and Rector-Brooks, Jarrid and Wolf, Guy and Bengio, Yoshua},
	month = mar,
	year = {2024},
	note = {arXiv:2302.00482 [cs]},
}

@misc{rombach_high-resolution_2022,
	title = {High-{Resolution} {Image} {Synthesis} with {Latent} {Diffusion} {Models}},
	url = {http://arxiv.org/abs/2112.10752},
	doi = {10.48550/arXiv.2112.10752},
	urldate = {2026-02-17},
	publisher = {arXiv},
	author = {Rombach, Robin and Blattmann, Andreas and Lorenz, Dominik and Esser, Patrick and Ommer, Björn},
	month = apr,
	year = {2022},
	note = {arXiv:2112.10752 [cs]},
}

@misc{ulyanov_instance_2017,
	title = {Instance {Normalization}: {The} {Missing} {Ingredient} for {Fast} {Stylization}},
	shorttitle = {Instance {Normalization}},
	url = {http://arxiv.org/abs/1607.08022},
	doi = {10.48550/arXiv.1607.08022},
	urldate = {2026-02-17},
	publisher = {arXiv},
	author = {Ulyanov, Dmitry and Vedaldi, Andrea and Lempitsky, Victor},
	month = nov,
	year = {2017},
	note = {arXiv:1607.08022 [cs]},
}

@misc{li_back_2026,
	title = {Back to {Basics}: {Let} {Denoising} {Generative} {Models} {Denoise}},
	shorttitle = {Back to {Basics}},
	url = {http://arxiv.org/abs/2511.13720},
	doi = {10.48550/arXiv.2511.13720},
	urldate = {2026-02-17},
	publisher = {arXiv},
	author = {Li, Tianhong and He, Kaiming},
	month = jan,
	year = {2026},
	note = {arXiv:2511.13720 [cs]},
}

@misc{ho_video_2022,
	title = {Video {Diffusion} {Models}},
	url = {http://arxiv.org/abs/2204.03458},
	doi = {10.48550/arXiv.2204.03458},
	urldate = {2026-02-17},
	publisher = {arXiv},
	author = {Ho, Jonathan and Salimans, Tim and Gritsenko, Alexey and Chan, William and Norouzi, Mohammad and Fleet, David J.},
	month = jun,
	year = {2022},
	note = {arXiv:2204.03458 [cs]},
}

@misc{wan_wan_2025,
	title = {Wan: {Open} and {Advanced} {Large}-{Scale} {Video} {Generative} {Models}},
	shorttitle = {Wan},
	url = {http://arxiv.org/abs/2503.20314},
	doi = {10.48550/arXiv.2503.20314},
	urldate = {2026-02-17},
	publisher = {arXiv},
	author = {Wan, Team and Wang, Ang and Ai, Baole and Wen, Bin and Mao, Chaojie and Xie, Chen-Wei and Chen, Di and Yu, Feiwu and Zhao, Haiming and Yang, Jianxiao and Zeng, Jianyuan and Wang, Jiayu and Zhang, Jingfeng and Zhou, Jingren and Wang, Jinkai and Chen, Jixuan and Zhu, Kai and Zhao, Kang and Yan, Keyu and Huang, Lianghua and Feng, Mengyang and Zhang, Ningyi and Li, Pandeng and Wu, Pingyu and Chu, Ruihang and Feng, Ruili and Zhang, Shiwei and Sun, Siyang and Fang, Tao and Wang, Tianxing and Gui, Tianyi and Weng, Tingyu and Shen, Tong and Lin, Wei and Wang, Wei and Wang, Wei and Zhou, Wenmeng and Wang, Wente and Shen, Wenting and Yu, Wenyuan and Shi, Xianzhong and Huang, Xiaoming and Xu, Xin and Kou, Yan and Lv, Yangyu and Li, Yifei and Liu, Yijing and Wang, Yiming and Zhang, Yingya and Huang, Yitong and Li, Yong and Wu, You and Liu, Yu and Pan, Yulin and Zheng, Yun and Hong, Yuntao and Shi, Yupeng and Feng, Yutong and Jiang, Zeyinzi and Han, Zhen and Wu, Zhi-Fan and Liu, Ziyu},
	month = apr,
	year = {2025},
	note = {arXiv:2503.20314 [cs]},
}

@misc{he_latent_2023,
	title = {Latent {Video} {Diffusion} {Models} for {High}-{Fidelity} {Long} {Video} {Generation}},
	url = {http://arxiv.org/abs/2211.13221},
	doi = {10.48550/arXiv.2211.13221},
	urldate = {2026-02-17},
	publisher = {arXiv},
	author = {He, Yingqing and Yang, Tianyu and Zhang, Yong and Shan, Ying and Chen, Qifeng},
	month = mar,
	year = {2023},
	note = {arXiv:2211.13221 [cs]},
}

@article{price_probabilistic_2025,
	title = {Probabilistic weather forecasting with machine learning},
	volume = {637},
	issn = {0028-0836},
	number = {8044},
	journal = {Nature},
	publisher = {Nature Publishing Group UK London},
	author = {Price, Ilan and Sanchez-Gonzalez, Alvaro and Alet, Ferran and Andersson, Tom R and El-Kadi, Andrew and Masters, Dominic and Ewalds, Timo and Stott, Jacklynn and Mohamed, Shakir and Battaglia, Peter},
	year = {2025},
	pages = {84--90},
}

@article{schwerdtner_uncertainty_2024,
	title = {Uncertainty quantification in coupled wildfire–atmosphere simulations at scale},
	volume = {3},
	issn = {2752-6542},
	number = {12},
	journal = {PNAS nexus},
	publisher = {Oxford University Press US},
	author = {Schwerdtner, Paul and Law, Frederick and Wang, Qing and Gazen, Cenk and Chen, Yi-Fan and Ihme, Matthias and Peherstorfer, Benjamin},
	year = {2024},
	pages = {pgae554},
}

@article{yu_probabilistic_2025,
	title = {A probabilistic approach to wildfire spread prediction using a denoising diffusion surrogate model},
	journal = {arXiv preprint arXiv:2507.00761},
	author = {Yu, Wenbo and Ghosh, Anirbit and Finn, Tobias Sebastian and Arcucci, Rossella and Bocquet, Marc and Cheng, Sibo},
	year = {2025},
}

@misc{kovachki_neural_2024,
	title = {Neural {Operator}: {Learning} {Maps} {Between} {Function} {Spaces}},
	shorttitle = {Neural {Operator}},
	url = {http://arxiv.org/abs/2108.08481},
	doi = {10.5555/3648699.3648788},
	urldate = {2026-02-18},
	author = {Kovachki, Nikola and Li, Zongyi and Liu, Burigede and Azizzadenesheli, Kamyar and Bhattacharya, Kaushik and Stuart, Andrew and Anandkumar, Anima},
	month = may,
	year = {2024},
	note = {arXiv:2108.08481 [cs]},
}

@misc{zou_enforcing_2026,
	title = {Enforcing {Reciprocity} in {Operator} {Learning} for {Seismic} {Wave} {Propagation}},
	url = {http://arxiv.org/abs/2602.11631},
	doi = {10.48550/arXiv.2602.11631},
	urldate = {2026-02-18},
	publisher = {arXiv},
	author = {Zou, Caifeng and Shi, Yaozhong and Ross, Zachary E. and Clayton, Robert W. and Azizzadenesheli, Kamyar},
	month = feb,
	year = {2026},
	note = {arXiv:2602.11631 [physics]},
}

@misc{wen_geometry_2026,
	title = {Geometry {Aware} {Operator} {Transformer} as an {Efficient} and {Accurate} {Neural} {Surrogate} for {PDEs} on {Arbitrary} {Domains}},
	url = {http://arxiv.org/abs/2505.18781},
	doi = {10.48550/arXiv.2505.18781},
	urldate = {2026-02-18},
	publisher = {arXiv},
	author = {Wen, Shizheng and Kumbhat, Arsh and Lingsch, Levi and Mousavi, Sepehr and Zhao, Yizhou and Chandrashekar, Praveen and Mishra, Siddhartha},
	month = jan,
	year = {2026},
	note = {arXiv:2505.18781 [cs]},
}

@misc{wu_transolver_2024,
	title = {Transolver: {A} {Fast} {Transformer} {Solver} for {PDEs} on {General} {Geometries}},
	shorttitle = {Transolver},
	url = {http://arxiv.org/abs/2402.02366},
	doi = {10.48550/arXiv.2402.02366},
	urldate = {2026-02-18},
	publisher = {arXiv},
	author = {Wu, Haixu and Luo, Huakun and Wang, Haowen and Wang, Jianmin and Long, Mingsheng},
	month = jun,
	year = {2024},
	note = {arXiv:2402.02366 [cs]},
}

@misc{alkin_universal_2025,
	title = {Universal {Physics} {Transformers}: {A} {Framework} {For} {Efficiently} {Scaling} {Neural} {Operators}},
	shorttitle = {Universal {Physics} {Transformers}},
	url = {http://arxiv.org/abs/2402.12365},
	doi = {10.48550/arXiv.2402.12365},
	urldate = {2026-02-18},
	publisher = {arXiv},
	author = {Alkin, Benedikt and Fürst, Andreas and Schmid, Simon and Gruber, Lukas and Holzleitner, Markus and Brandstetter, Johannes},
	month = feb,
	year = {2025},
	note = {arXiv:2402.12365 [cs]},
}

@article{yang_seismic_2021,
	title = {Seismic {Wave} {Propagation} and {Inversion} with {Neural} {Operators}},
	volume = {1},
	issn = {2694-4006},
	url = {https://doi.org/10.1785/0320210026},
	doi = {10.1785/0320210026},
	number = {3},
	urldate = {2026-02-18},
	journal = {The Seismic Record},
	author = {Yang, Yan and Gao, Angela F. and Castellanos, Jorge C. and Ross, Zachary E. and Azizzadenesheli, Kamyar and Clayton, Robert W.},
	month = nov,
	year = {2021},
	pages = {126--134},
}

@article{zou_deep_2024,
	title = {Deep neural {Helmholtz} operators for 3-{D} elastic wave propagation and inversion},
	volume = {239},
	issn = {1365-246X},
	url = {https://doi.org/10.1093/gji/ggae342},
	doi = {10.1093/gji/ggae342},
	number = {3},
	urldate = {2026-02-18},
	journal = {Geophysical Journal International},
	author = {Zou, Caifeng and Azizzadenesheli, Kamyar and Ross, Zachary E and Clayton, Robert W},
	month = dec,
	year = {2024},
	pages = {1469--1484},
}

@misc{shi_universal_2024,
	title = {Universal {Functional} {Regression} with {Neural} {Operator} {Flows}},
	url = {http://arxiv.org/abs/2404.02986},
	doi = {10.48550/arXiv.2404.02986},
	urldate = {2026-02-18},
	publisher = {arXiv},
	author = {Shi, Yaozhong and Gao, Angela F. and Ross, Zachary E. and Azizzadenesheli, Kamyar},
	month = nov,
	year = {2024},
	note = {arXiv:2404.02986 [cs]},
}

@misc{yao_guided_2026,
	title = {Guided {Diffusion} {Sampling} on {Function} {Spaces} with {Applications} to {PDEs}},
	url = {http://arxiv.org/abs/2505.17004},
	doi = {10.48550/arXiv.2505.17004},
	urldate = {2026-02-18},
	publisher = {arXiv},
	author = {Yao, Jiachen and Mammadov, Abbas and Berner, Julius and Kerrigan, Gavin and Ye, Jong Chul and Azizzadenesheli, Kamyar and Anandkumar, Anima},
	month = feb,
	year = {2026},
	note = {arXiv:2505.17004 [cs]},
}

@misc{lipman_flow_2023,
	title = {Flow {Matching} for {Generative} {Modeling}},
	url = {http://arxiv.org/abs/2210.02747},
	doi = {10.48550/arXiv.2210.02747},
	urldate = {2026-02-18},
	publisher = {arXiv},
	author = {Lipman, Yaron and Chen, Ricky T. Q. and Ben-Hamu, Heli and Nickel, Maximilian and Le, Matt},
	month = feb,
	year = {2023},
	note = {arXiv:2210.02747 [cs]},
}

@article{lavrentiadis_seismologically_2025,
	title = {A seismologically consistent surface rupture length model for unbounded and width-limited events},
	volume = {41},
	issn = {8755-2930},
	number = {4},
	journal = {Earthquake Spectra},
	publisher = {SAGE Publications Sage UK: London, England},
	author = {Lavrentiadis, Grigorios and Wang, Yongfei and Abrahamson, Norman A and Bozorgnia, Yousef and Goulet, Christine A},
	year = {2025},
	pages = {2859--2879},
}

@article{miller_groundmotion_2015,
	title = {Ground‐motion intensity and damage map selection for probabilistic infrastructure network risk assessment using optimization},
	volume = {44},
	issn = {0098-8847},
	number = {7},
	journal = {Earthquake Engineering \& Structural Dynamics},
	publisher = {Wiley Online Library},
	author = {Miller, M and Baker, J},
	year = {2015},
	pages = {1139--1156},
}

@book{baker_seismic_2021,
	title = {Seismic hazard and risk analysis},
	isbn = {1-108-60490-0},
	publisher = {Cambridge University Press},
	author = {Baker, Jack and Bradley, Brendon and Stafford, Peter},
	year = {2021},
}

\newpage

\textbf{Acknowledgments:}We thank Arben Pitarka for helpful discussions regarding the preparation of the finite-rupture simulations, and Flora Xia for valuable input on the validation metrics. \textbf{Funding:}
This material is based upon work supported by the U.S. Department of Energy, Office of Science, Office of Advanced Scientific Computing Research, Science Foundations for Energy Earthshot under Award Number DE-SC0024705. ZER is supported by a fellowship from the David and Lucile Packard Foundation. \textbf{Author contributions:} Conceptualization: D.A., Y.S., and Z.R.E., Methodology : Y.S., Z.R.E., D.A., G.L., K.Z., and C.Z. Dataset Preparation : K.T., D.M., and G.L., Writing : Y.S., G.L., K.T., Z.E.R., and D.A., Visualziation : Y.S., G.L., K.T., and D.A., \textbf{Competing interests:} The authors declare that they have no competing interests.
\textbf{Data, code and materials availablility} All materials required to reproduce the results in the paper are publicly available.

\newpage
\appendix
\setcounter{figure}{0}
\setcounter{table}{0}
\renewcommand{\thefigure}{S\arabic{figure}}
\renewcommand{\thetable}{S\arabic{table}}

\appendix
\begin{center}
    {\LARGE \textbf{Supplementary Materials}\par}
    \vspace{1cm} 
\end{center}

\section{Background of Neural Operator and Flow Matching}

\textbf{Neural Operators.} Neural operators are defined in infinite-dimensional function space, and generalize traditional finite-dimensional neural network~\cite{kovachki_neural_2024, azizzadenesheli_neural_2024}. Different from neural works, neural operators  are often described as discretization-agnostic, which treat inputs and outputs as functions with arbitrary discretizations. In seismology, neural operators have demonstrated remarkably progress in solving the wave equation, delivering orders-of-magnitude speedups over conventional numerical solvers~\cite{yang_seismic_2021, zou_deep_2024, aquib_broadband_2024} while enabling the enforcement of explicit physical constraints~\cite{zou_enforcing_2026}. Recent advances in operator learning aim to develop more expressive and scalable neural operator architectures that works irregular grids~\cite{alkin_universal_2025, shi_mesh-informed_2025, wen_geometry_2026, wu_transolver_2024}. Another important direction is to develop functional generative model via neural operators, with application in modeling stochastic processes~\cite{shi_universal_2024, shi_stochastic_2025} and for probabilistic PDE solving~\cite{yao_guided_2026}.

\textbf{Flow Matching.} Flow matching is a special continuous normalizing flow framework, learns a time-dependent velocity field that transports samples from a simple reference distribution (e.g., Gaussian distribution) to the data distribution via a simulation-free regression objective~\cite{lipman_flow_2023}. Especially, rectified flow or equivalently, independent coupling optimal-transport flow matching are widely used due to their concise mathematical formulation, robust performance, and efficiency in training and inference~\cite{liu_flow_2022, tong_improving_2024}. Finally, recent work revisits the manifold assumption and finds that, directly predicting the clean sample is more tractable than predicting off-manifold quantities such as noise or flow velocity which suggests to train the flow matching model to directly output a clean estimate, then deterministically convert clean prediction into the corresponding velocity (or noise) using the predefined linear interpolation scheduler. The "clean-prediction, velocity loss" yields a formulation that keeps the inductive bias of clean prediction while retaining a well-behaved velocity-matching objective and favorable effective weighting in practice.~\cite{li_back_2026}.

\section{Details of simulation dataset preparation}

\begin{figure*}[ht]
    \vspace*{-.3cm}
    \centering

    \includegraphics[width=0.95\textwidth,trim={1cm, 1.5cm, 1cm, 1.5cm},clip]
    {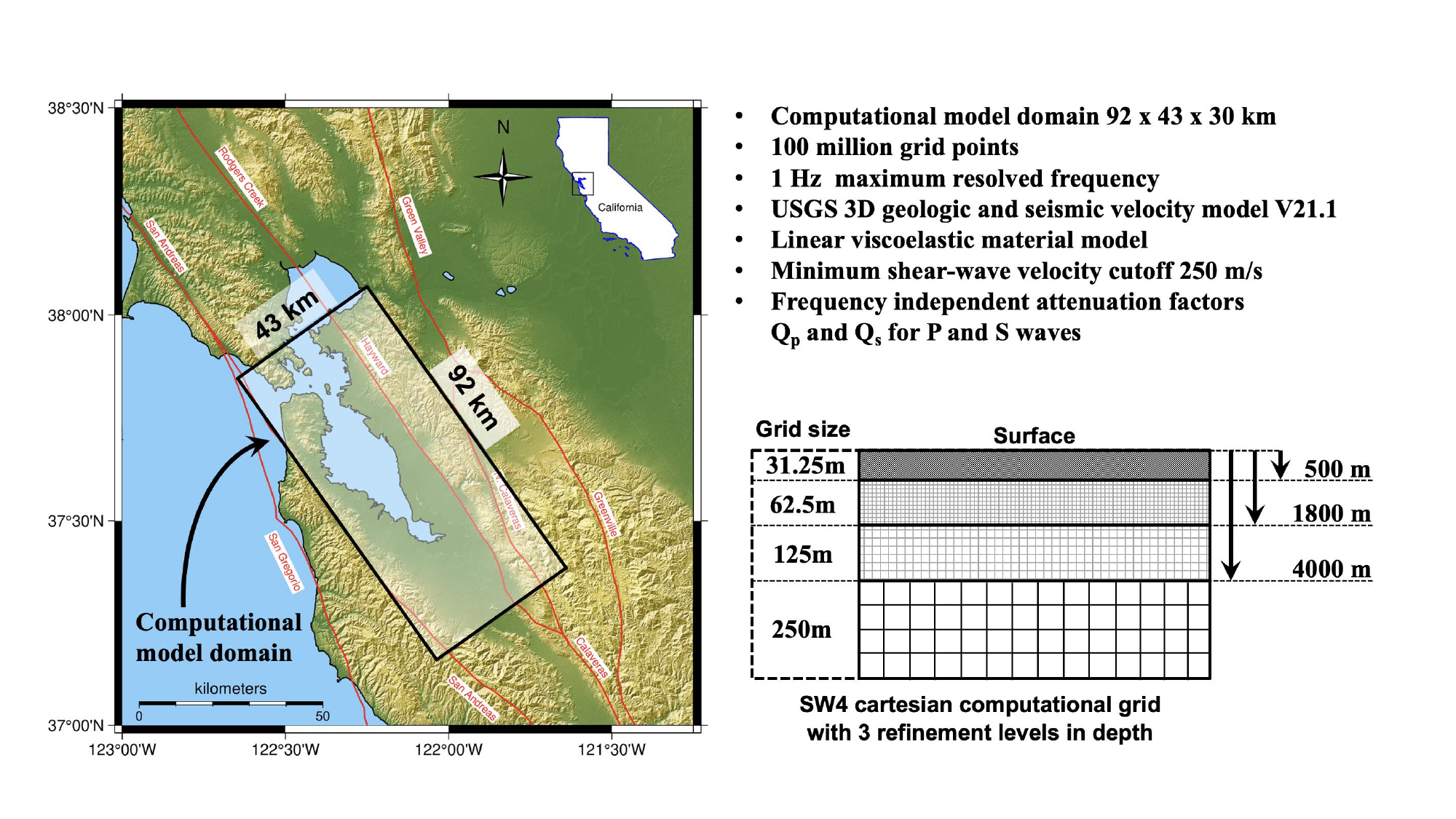}
\caption{Computational domain adopted for the SFBA and model parameters}
  \label{fig:supp_fig_s1}   
\end{figure*}

\textbf{Computational testbed.} Selecting an appropriate computational testbed is critical to ensuring the ability of the AI models to capture potential complexities of realistic wave propagation. It is thus emphasized that the simulation domain must exhibit heterogeneous geological features such as shallow sedimentary basins and sharp contrasts between different geologic units that result in abrupt changes in seismic wave velocity propagation and strongly influence spatial amplification patterns. The San Francisco Bay Area (SFBA) satisfies these criteria (Fig~\ref{fig:supp_fig_s1}), and additionally is a densely populated metropolitan region with significant infrastructure development and seismic activity resulting in a potential of high risk, thus has been significantly studied. For this study, we adopted the United States Geological Survey (USGS) SFBA community velocity model (version 21.1)~\cite{aagaard_san_2021}. This model was selected because it has undergone rigorous performance evaluation through high-resolution 3D physics-based simulations and recorded seismicity. A recent validation study ~\cite{pinilla-ramos_performance_2025} in the SFBA involved small-magnitude earthquakes and demonstrated that the velocity model performs particularly well in the frequency range of 0.25–3 Hz with practically unbiased simulated motions compared to recorded ground motion data and slight over-prediction of 25$\%$ in the frequency range 3-5Hz, successfully capturing wave propagation effects in near-surface soft materials and local basins. Moreover, using the same testbed, a significant simulation effort of large-magnitude Hayward-fault earthquake events has been recently conducted as a collaboration between the Pacific Earthquake Engineering Research (PEER) Center and Lawrence Berkeley National Laboratory (LBNL)~\cite{mccallen_open-access_2025}. This work leverages the EQSIM framework and Exascale computing for earthquake simulation capabilities~\cite{mccallen_regional-scale_2024, mccallen_eqsimmultidisciplinary_2021}, and produced an open-access Simulated Ground Motion Database (SGMD) of surface motions in the SFBA. 

\textbf{Physics-based earthquake simulations.}
The physics-based ground motion simulations in this study are generated using the SW4 (Seismic Waves, 4th order) software. SW4 numerically solves the elastodynamic wave equations in displacement formulation using a node-based finite difference approach, guaranteeing energy stability and achieving fourth-order accuracy in both space and time. The primary objective of these simulations is to capture the deterministic physics of wave propagation, including the effects of complex 3D geologic structures such as sedimentary basins, which strongly influence ground motion amplitude and duration. While the simulations provide a rigorous representation of linear wave propagation, they are subject to limitations. First, the modeling assumes a linear-elastic response of the geologic materials, neglecting potential nonlinear soil behavior. Second, simulated high-frequency content is limited by epistemic uncertainty of subsurface conditions, i.e., the lack of sufficient data on small-scale soil properties (on the order of tens of meters) to construct a velocity model capable of supporting accurate wave propagation at frequencies above 3-5 Hz. Simultaneously, a minimum shear-wave velocity cutoff (e.g., 250 m/s adopted here) is enforced because lower velocities correspond to shorter wavelengths, necessitating a finer computational grid that would drastically increase computational demands. This constraint is critical because computational cost scales inversely with the fourth power of the grid spacing (which is driven by the ratio of frequency to minimum shear-wave velocity), making high-resolution deterministic simulations computationally intensive. Despite these limitations, physics-based simulations provide high-fidelity data for medium- to long-period wave propagation. For the dataset generation in this study, the simulations are designed to resolve frequencies up to 1 Hz aiming to achieve manageable computational effort and robust simulated datasets of verified validity for the physical domain of interest. The following subsections describe the adopted computational domain, the seismic source characterization for small- and large-magnitude earthquake simulations and the simulation strategy for efficiently generating the ground motion datasets.

\textbf{Computational domain and velocity model.}
The computational domain is shown in Fig~\ref{fig:supp_fig_s1} and is identical to the one used in~\cite{pinilla-ramos_performance_2025}which is a sub-domain of the detailed USGS v21.1 SFBA V~\cite{aagaard_san_2021}. The main modelling assumptions are listed in Figure 2, including a sketch describing the grid size and the implemented grid refinements in depth. A minimum shear-wave velocity of 250 m/s is adopted, and combined with a required accuracy for frequencies up to 1 Hz results in a minimum grid spacing of 31.25 m to ensure proper numerical accuracy ensuring 8 points per wavelength in all points within the mesh. The wave attenuation is modeled through a linear visco-elastic model characterized by the quality factors (Qp and Qs for compressional and shear waves, respectively), which were obtained directly from the characterization included in the velocity model by the USGS.

\begin{figure*}[ht]
    \vspace*{-.3cm}
    \centering

    \includegraphics[width=0.95\textwidth,trim={1cm, 0.5cm, 1.5cm, 0.5cm},clip]
    {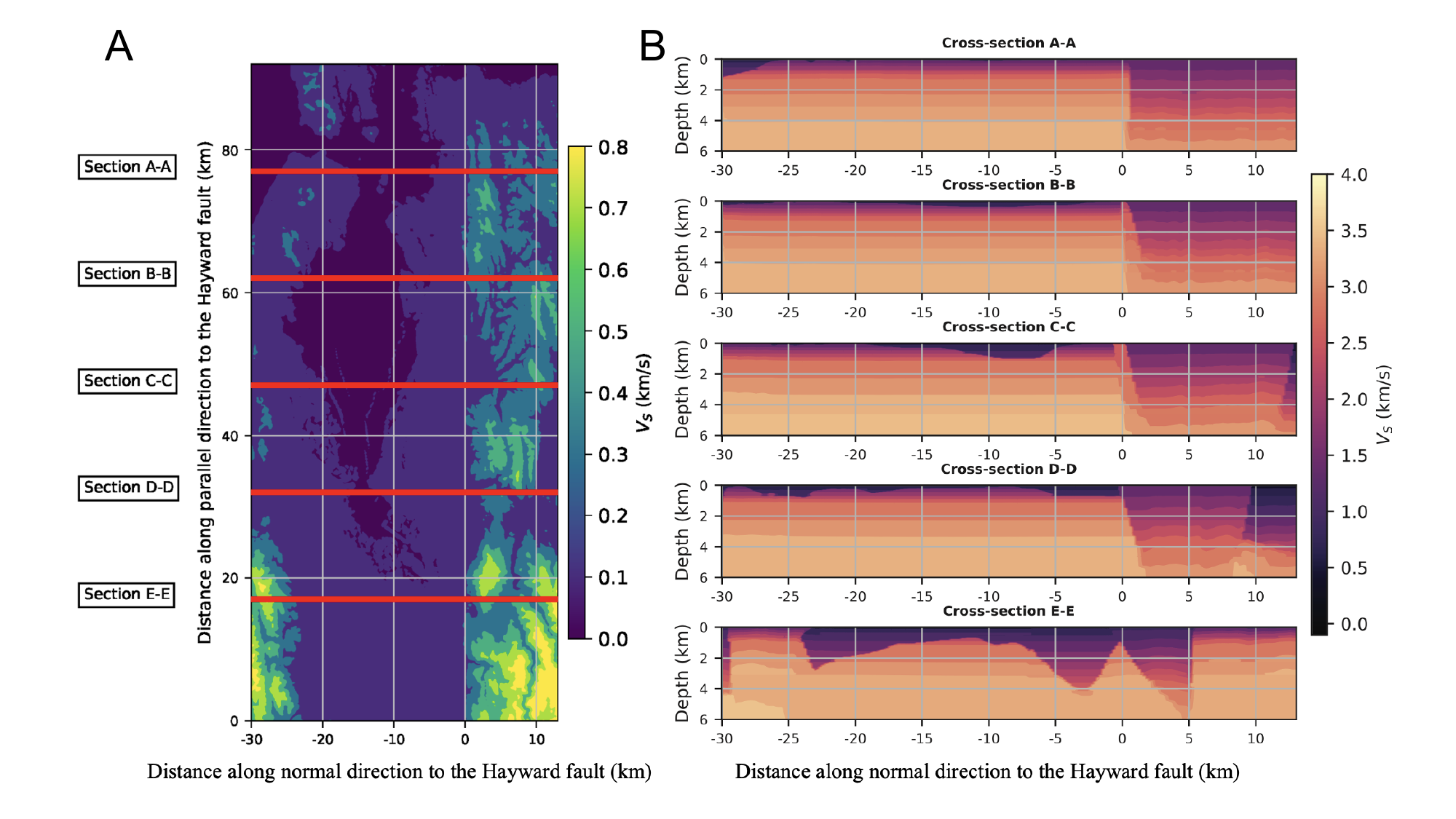}
\caption{(a) Map view of the free surface shear-wave velocity in the simulation domain, extracted from the United States Geological Survey velocity model, with the E-W red lines indicating the location of the velocity model cross-sections shown in (b); (b) Vertical cross-sections of the velocity model with the Hayward fault at 0 normal distance.
(adopted from Pinilla-Ramos et al. 2025 with permission)
}
  \label{fig:supp_fig_s2}   
\end{figure*}

The SFBA region features complex geologic structures that significantly influence wave propagation. In the northern area, the Hayward fault separates two distinct geological blocks~\cite{pinilla-ramos_performance_2025} (cross-sections A-A, B-B, and C-C in Fig~\ref{fig:supp_fig_s2}). The western block consists of shallow soft marine sediments overlying sedimentary basins that extend down to 1 km depth (cross-sections B-B and C-C). In contrast, the eastern block consists mainly of sedimentary rocks and tends to be stiffer at the surface (with shear-wave velocities usually above 500 m/s). This comparison reverses at depths larger than 1 km where east of the Hayward fault consists of softer rocks compared to the western block. Structural complexity increases to the south where the Calaveras fault bounds the eastern block (cross-sections C-C, D-D, and E-E). Notably, the San Jose and Santa Clara regions in the south of the SFBA (cross-section E-E) exhibit deep, irregular sedimentary basins, which are expected to generate significant wave reverberations and increase ground motion duration.

\begin{figure*}[ht]
    \vspace*{-.3cm}
    \centering

    \includegraphics[width=0.9\textwidth,trim={3cm, 0.5cm, 4.5cm, 0.5cm},clip]
    {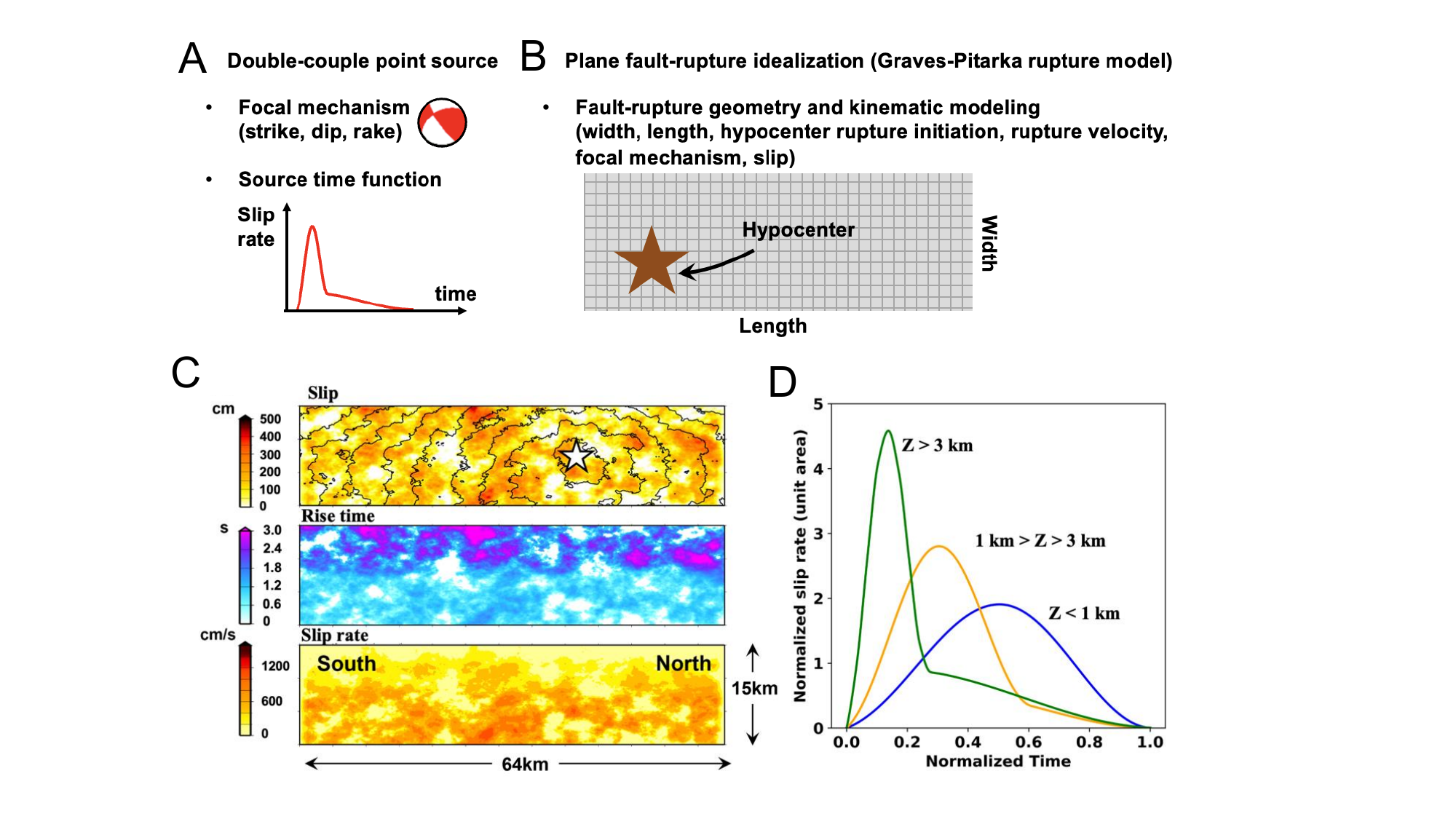}
\caption{(a) Small-magnitude seismic source characterization as a double-couple point-source function defined by its focal mechanism and source time function; (b) Large-magnitude seismic source characterization following the Graves-Pitarka rupture model; (c) Example of a pure stochastic rupture realization for the Hayward fault plane, shown through its total slip, rise time and slip rate ; (d) slip rate of points at various depths within the fault-plane as a function of time, normalized to show a unit area of total slip, developed within a normalized rise time of 1 second.}
  \label{fig:supp_fig_s3}   
\end{figure*}

\textbf{Seismic source characterization}
The simulation of small-magnitude earthquake events (i.e., magnitudes less than 5) are considered following a double-couple point source using the Liu source-time function~\cite{liu_prediction_2006} to model how the source releases energy over time. Being consistent with rupture dynamics, this slip rate function is not symmetric (as shown in Fig~\ref{fig:supp_fig_s3}; it is characterized by a large initial peak and a gradual amplitude decay that represents the healing process of the rupture.
To simulate large-magnitude earthquake events, a planar fault rupture is modelled and initiates (at the hypocenter location) evolving with time during the earthquake event. The rupture process for a given scenario is controlled by numerous seismological properties including the total slip, earthquake realizations utilized in the SFBA simulations are generated using the Graves and Pitarka (GP) kinematic rupture model~\cite{graves_kinematic_2016}, with an enhanced slip rate representation~\cite{pitarka_refinements_2022}.
The shape of the source time function varies with depth with a linear transition between crustal depths of 1 and 3km. This formulation allows the shape of the source time function to transition from “cosine-type,” with a relatively modest high-frequency content in the depth interval 0–1km, to “Kostrov-type” with a broadband frequency content, at depths greater than 3km.

\begin{figure*}[ht]
    \vspace*{-.3cm}
    \centering

    \includegraphics[width=0.9\textwidth,trim={1cm, 4cm, 1.5cm, 4cm},clip]
    {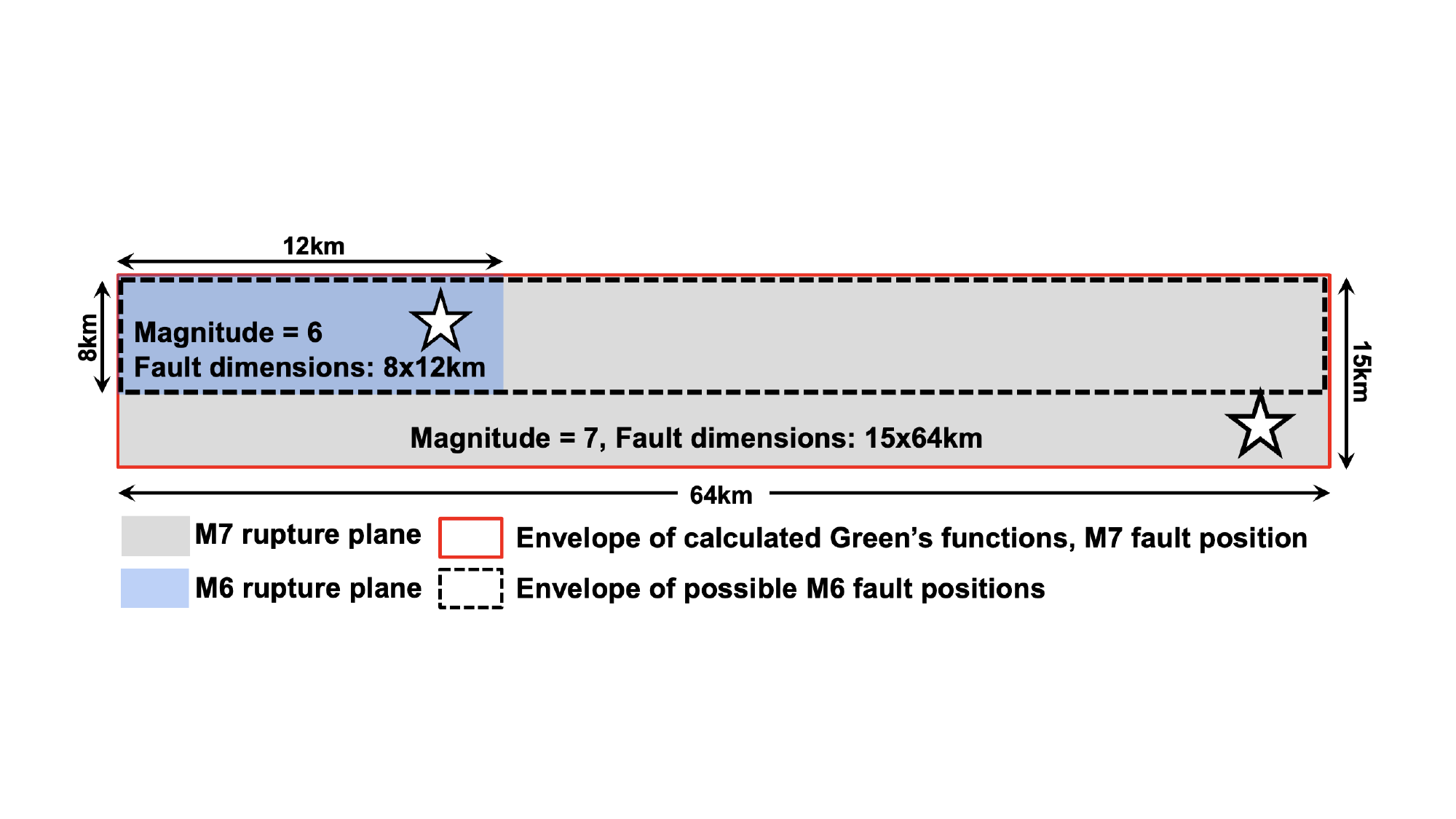}
\caption{Envelope of calculated Green’s functions, rupture plane dimensions and positions for Magnitude 7 and Magnitude 6 earthquake events}
  \label{fig:supp_fig_s4}   
\end{figure*}

\begin{figure*}[ht]
    \vspace*{-.3cm}
    \centering

    \includegraphics[width=0.9\textwidth,trim={1cm, 0.5cm, 1.5cm, 0.5cm},clip]
    {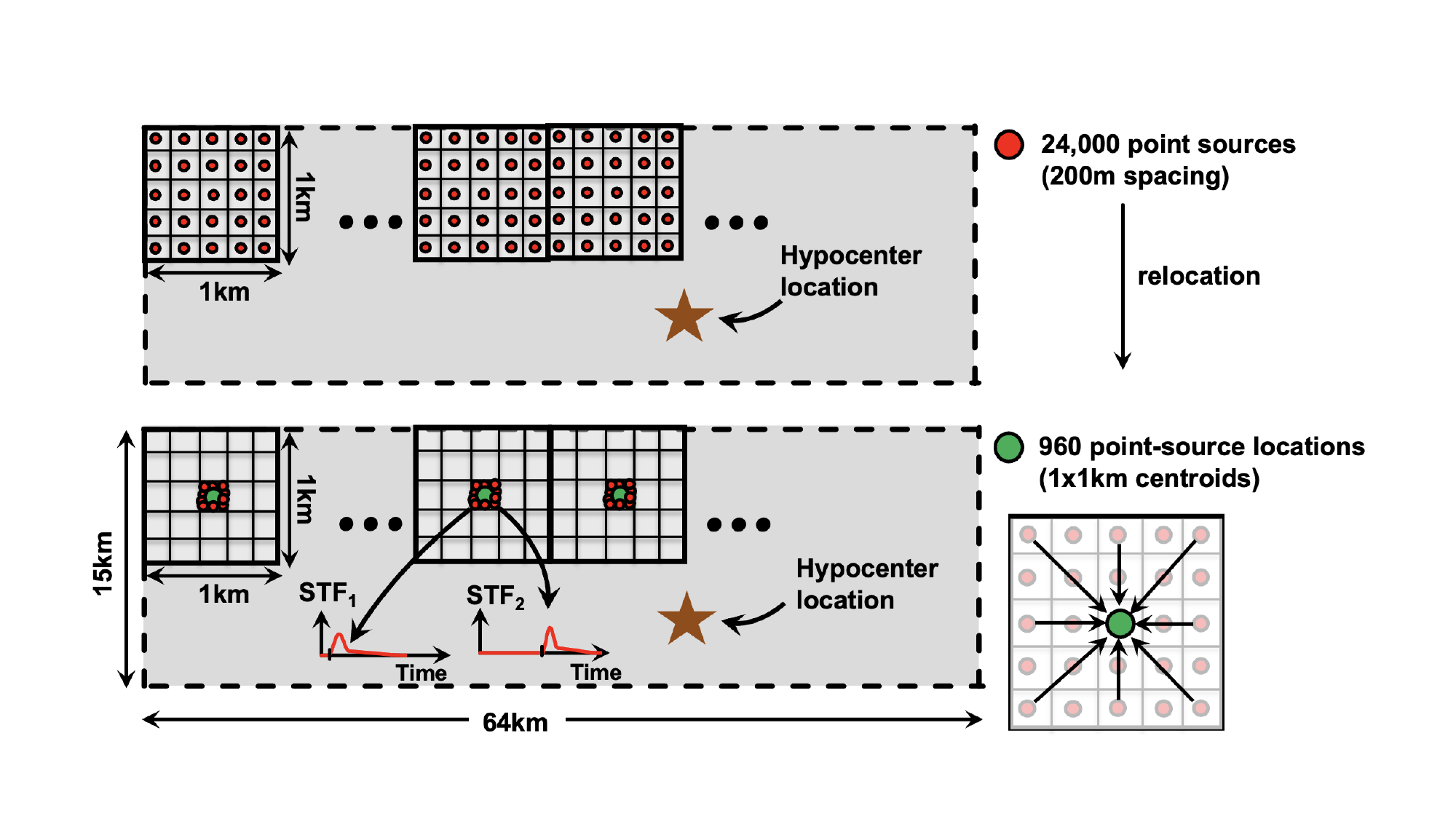}
\caption{Fault-plane discretization with point-sources at 200-m grid, repositioned in the corresponding centroids of 1x1km sub-faults}
  \label{fig:supp_fig_s5_rupture_conv}   
\end{figure*}

\textbf{Simulation database generation} The database contains 2,100 small-magnitude earthquake events, with 2,000 intended for training of the AI algorithms and 100 for testing. All simulations have the same focal mechanism of dip=80°, strike=145° and rake=180° which has been identified as the one in the 2018 Berkeley 4.4-magnitude earthquake event reported in~\citep{pinilla-ramos_performance_2025}, as well as the Liu source function and corner frequency of fc=1.4Hz suggested in the same study. The locations of the small-magnitude earthquake simulations are randomly positioned throughout the domain, covering its entire width, length and depth using the Latin Hypercube Sampling technique to ensure uniform spatial distribution. The simulations are performed on Perlmutter, a 70-PF (Peta flops) Graphics Processing Unit (GPU)-accelerated super-computer at the National Energy Research Super Computer Center at Lawrence Berkeley National Laboratory, with a total of 600 node-hours of GPU resources, where each node contains 4 NVIDIA A100 GPUs. Moreover, the database contains large-magnitude earthquake simulation, in particular 1,100 M7 and 2,100 M6 earthquake realizations, out of which 100 M7 and 100 M6 realizations are reserved for testing and the rest are reserved for training. To generate the source ruptures for each of those realizations, the Graves-Pitarka kinematic rupture model is deployed ~\cite{graves_kinematic_2016}.The rupture plane dimensions are shown in Figure 6, and are governed by the magnitude of the seismic event, in particular 15x64 km for the M7 and 8x12km for the M6.  
 
The straightforward approach in EQSIM~\cite{mccallen_eqsimmultidisciplinary_2021-1} is to directly incorporate the extended source rupture files into sw4 and simulate the earthquake response. However, this could become cumbersome for generating thousands of realizations, and thus a different approach is investigated here to ensure low computational cost as well as scalability (i.e., the ability to generate more realizations if needed without the associated scaling of the computational cost). To this end, the representation theorem is deployed, allowing to use the Green’s functions describing the source-to-site response (i.e., from each point of the finite source rupture to the surface stations).  In Fig~\ref{fig:supp_fig_s4}, the envelope of the plane in which the Green’s functions are calculated is shown in red, which corresponds to the M7 rupture plane shown as a gray-shaded area. The M6 rupture plane is shown as a blue-shaded area, and given its smaller dimensions compared to the M7 rupture plane, it can be positioned anywhere within the black-dashed envelope (i.e., as a sliding stencil at the top of the envelope where the Green’s functions are calculated). For the M6 and M7 modelled magnitudes, the rupture variable parameters in these realizations are: 1) the hypocenter locations within the fault (shown in star markers in Fig~\ref{fig:supp_fig_s4}); 2) the slip distribution controlled by a random seed generator within the Graves-Pitarka kinematic rupture generator; and 3) the average rupture velocity . In particular, the hypocenter horizontal location (HH) is –25 km < HH < 25 km measured from the center of the M7 fault plane and –3 km < HH < 3 km measured from the center of the M6 fault plane. The hypocenter depth (HD) is 7 km < HD < 9 km measured from the top of the M7 fault plane and 3 km < HD < 6 km measured from the top of the M6 fault plane. The rupture velocity (VRup) is 0.65 < VRup < 0.95, expressed as a percentage of the average shear-wave velocity, similarly for both M7 and M6 ruptures. 

In Fig~\ref{fig:supp_fig_s5_rupture_conv}, the first row shows the 200-meter-spacing discretization of the rupture plane, as required for a proper rupture characterization of frequencies up to 1Hz, resulting in 24,000 point-sources. To reduce the volume of Green’s functions calculations, these point sources are then grouped into 5x5km sub-faults, and artificially considered to be located in their respective centroid that is shown in green at the second row of Figure 7. It is noted that during this modification, the respective source-time functions are preserved at the original resolution (24,000) to maintain the rupture model’s fidelity, and only the calculated Green’s functions are reduced to 960 (i.e., by a factor of 25). This allows for a significant reduction in computational demand – i.e., calculating 960 as opposed to 24,000 Green’s functions – requiring a total of 400 node-hours of GPU nodes on Perlmutter. After this dataset of Green’s functions is calculated, each realization defined by its source-rupture is calculated by convolving the 24,000 source time functions with their respective centroids, which additionally requires 1 node-hour of CPU-only nodes on Perlmutter, resulting in a total of 3,100 CPU-only node-hours, where each CPU-node contains two AMD EPYC 7763 (Milan) CPUs. 

\textbf{Preprocessing of the simulation data.}
To simplify the learning setting and alleviate artifacts induced by absorbing boundaries, we crop the simulations to an $80 \times 40$ km subdomain sharing the same geometric center as the original computational domain ($92 \times 43$). We further apply a sixth-order low-pass filter with cutoff frequency $1$ Hz, downsample the spatial discretization from $[320,160]$ to $[256,128]$, and truncate the time series to $24$ seconds sampled at (original) $4$ Hz. The resulting three-component velocity wavefields therefore have shape $[3,256,128,96]$. To alleviate dynamic-range variation across events of different magnitudes, each sample is divided by its sample-wise standard deviation, and the corresponding $\log_{10}$ standard deviation is appended as an additional constant channel. Consequently, each event is represented as an array of shape $[4,256,128,96]$. The unnormalized three-channel representation and the normalized four-channel representation are fully equivalent, since either can be recovered from the other.

\section{Details of experimental setup}

This section describes the experimental configuration and training protocol. All experiments were conducted on a single NVIDIA RTX A6000 Ada GPU (48~GB). All three components, namely the super-resolution neural operator (SNO), the autoencoding neural operator (AENO), and the conditional latent flow matching model, were trained on a total of 5{,}000 samples. Among them, only the latent flow matching model is stochastic and conditioned on event (magnitude, hypocenter), whereas SNO and AENO are deterministic operators.

\textbf{Super-resolution neural operator (SNO).}
The SNO maps a low-resolution wavefield to a high-resolution wavefield. During training, the low-resolution input has shape $[4,128,96,48]$ and the high-resolution target has shape $[4,256,128,96]$. The fourth channel is a normalization channel that stores the per-sample standard deviation on a $\log_{10}$ scale. We implement SNO using a residual Fourier neural operator (rFNO) with 4 FNO layers. The model has 871~million parameters and a storage. We train it for 100 epochs with batch size 2 using AdamW and a cosine-annealing learning-rate schedule with an initial learning rate of $5\times 10^{-4}$. The total training time is 58.5~hours.

\textbf{Autoencoding neural operator (AENO).}
The AENO consists of an encoder operator and a decoder operator, each implemented as an rFNO with 6 layers. The model has 663~million parameters. We train it for 200 epochs with batch size 8 using AdamW and a cosine-annealing learning-rate schedule with an initial learning rate of $5\times 10^{-4}$. The total training time is 36.3~hours.

\textbf{Conditional latent flow matching.}
For conditional latent flow matching, we adopt a diffusion U-Net architecture following~\cite{tong_improving_2024}. The model has 128~million parameters. We train it for 300 epochs with batch size 64 using AdamW and a cosine-annealing learning-rate schedule with an initial learning rate of $2\times 10^{-4}$. The total training time is 5.8~hours.

\begin{table}[h]
\centering
\caption{Hyperparameters and training details of the SNO, AENO, and Flow Matching models.}
\label{tab:experiment_details}
\begin{tabular}{l c c c}
\toprule
\textbf{Components} & \textbf{SNO} & \textbf{AENO} & \textbf{Flow Matching} \\
\midrule
Architecture & rFNO  & rFNO & Diffusion U-Net \\
Layers & 4 & 6+6 & - \\
Batch Size & 2 & 8 & 64 \\
Epochs & 100 & 200 & 300 \\
Learning Rate & $5 \times 10^{-4}$ & $5 \times 10^{-4}$ & $2 \times 10^{-4}$ \\
Parameters & 871 M & 663 M & 128 M \\
Learning rate & 5e-4 & 5e-4 & 2e-4 \\
Scheduler & Cosine Annealing & Cosine Annealing & Cosine Annealing  \\
Training Time & 58.5 h & 36.3 h & 5.8 h \\
\bottomrule
\end{tabular}
\end{table}

\section{Additional Results}

\textbf{Random reference stations for NCC.}
We provide additional NCC analyses using randomly selected reference stations for the representative \(\text{Mw}~4.4\) point-source scenario shown in Fig.~\ref{fig:point_source}. These supplementary examples are intended to verify that the coherence patterns reported in the main text are not specific to a particular reference location. Across different choices of reference station, the synthetic wavefields reproduce similar spatial patterns of peak NCC and, more importantly, similar travel-time gradients in the time-lag fields relative to the ground truth. This further supports the conclusion that \alg\ preserves physically consistent phase alignment and wave-propagation kinematics over the full spatial domain.

\textbf{Evaluation on the \(\text{Mw}~6.0\) finite-rupture source scenario.}
We also include additional evaluation results for the \(\text{Mw}~6.0\) finite-rupture scenario in Fig.~\ref{fig:extended_source_M6}. This case serves as an intermediate regime between the comparatively simple \(\text{Mw}~4.4\) point-source events and the more challenging \(\text{Mw}~7.0\) extended-rupture events discussed in the main text. The supplementary results show that \alg\ captures the principal characteristics of finite-rupture ground motion, including spatially varying amplitude patterns, coherent phase delays, and rupture-driven wavefield asymmetry. These results indicate that the model remains effective not only at the two ends of the magnitude range, but also in the intermediate finite-rupture regime.

\textbf{Zero-shot super-resolution on finer spatial grids.}
Finally, we present zero-shot super-resolution results in Fig.~\ref{fig:sup_res_supp}. In this experiment, the synthetic wavefields are evaluated on a finer spatial grid of \([3, 512, 256, 96]\), whereas the reference simulations are available at the default resolution of \([3, 256, 128, 96]\). Because \alg\ is formulated as a functional generative model with a super-resolution neural operator, the learned representation is not tied to a single output discretization. As a result, the model can be queried on denser spatial grids without retraining, while still producing wavefields that remain visually coherent and physically plausible.
\begin{figure*}[ht]
    \vspace*{-.3cm}
    \centering

    \includegraphics[width=0.95\textwidth,trim={1cm, 0, 1cm, 0},clip]
    {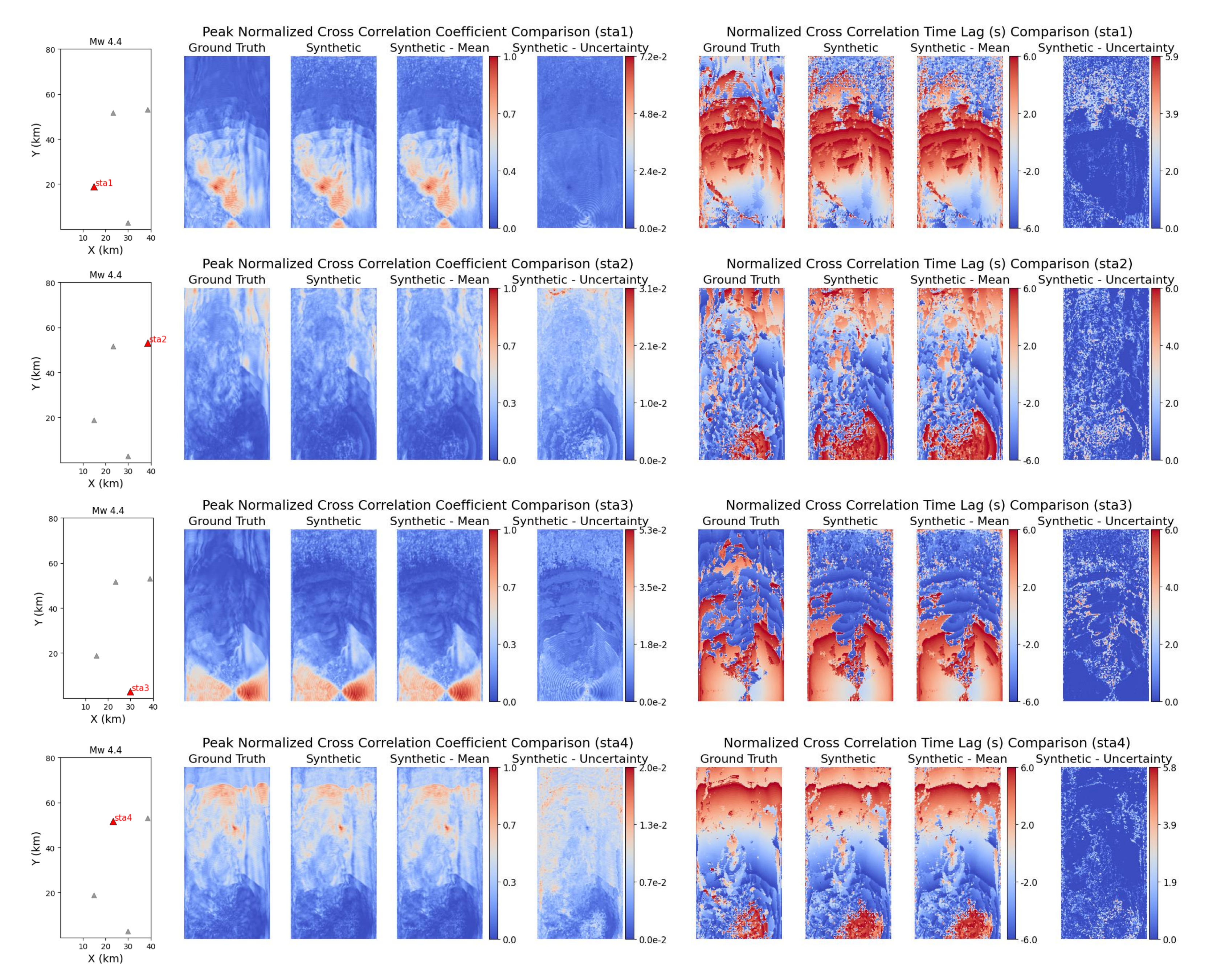}
\caption{NCC analysis using randomly selected reference stations for the representative $M_w~4.4$ event.}
  \label{fig:supp_fig_ncc}   
\end{figure*}

\begin{figure*}[ht]
    \vspace*{-.3cm}
    \centering

    \includegraphics[width=0.95\textwidth,trim={1cm, 0, 1cm, 0},clip]
    {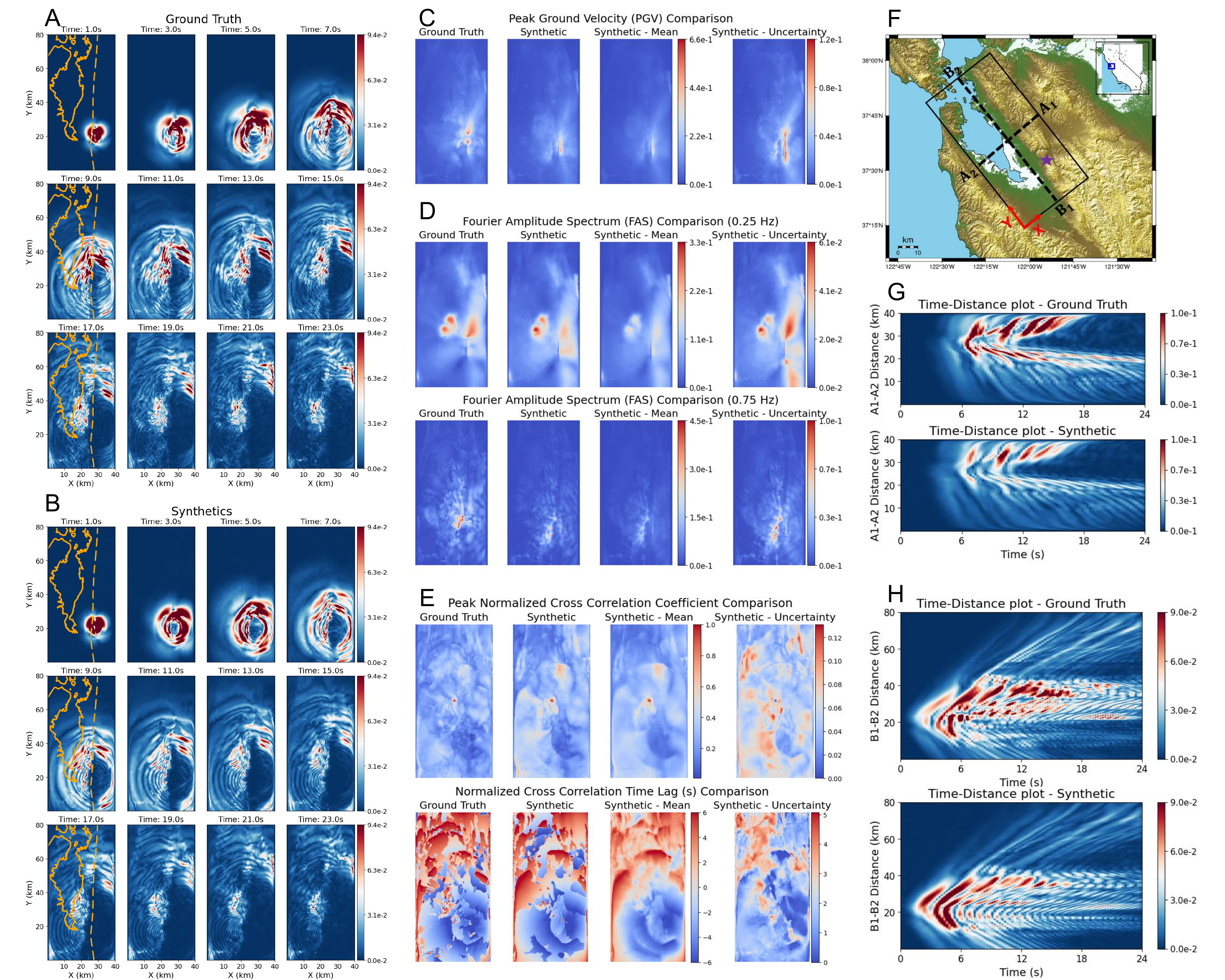}
    \caption{Validation of \alg on a representative $M_w~6.0$ finite rupture source scenario.}  \label{fig:extended_source_M6}   
\end{figure*}

\begin{figure*}[ht]
    \vspace*{-.3cm}
    \centering

    \includegraphics[width=0.95\textwidth,trim={1cm, 0, 2cm, 0},clip]
    {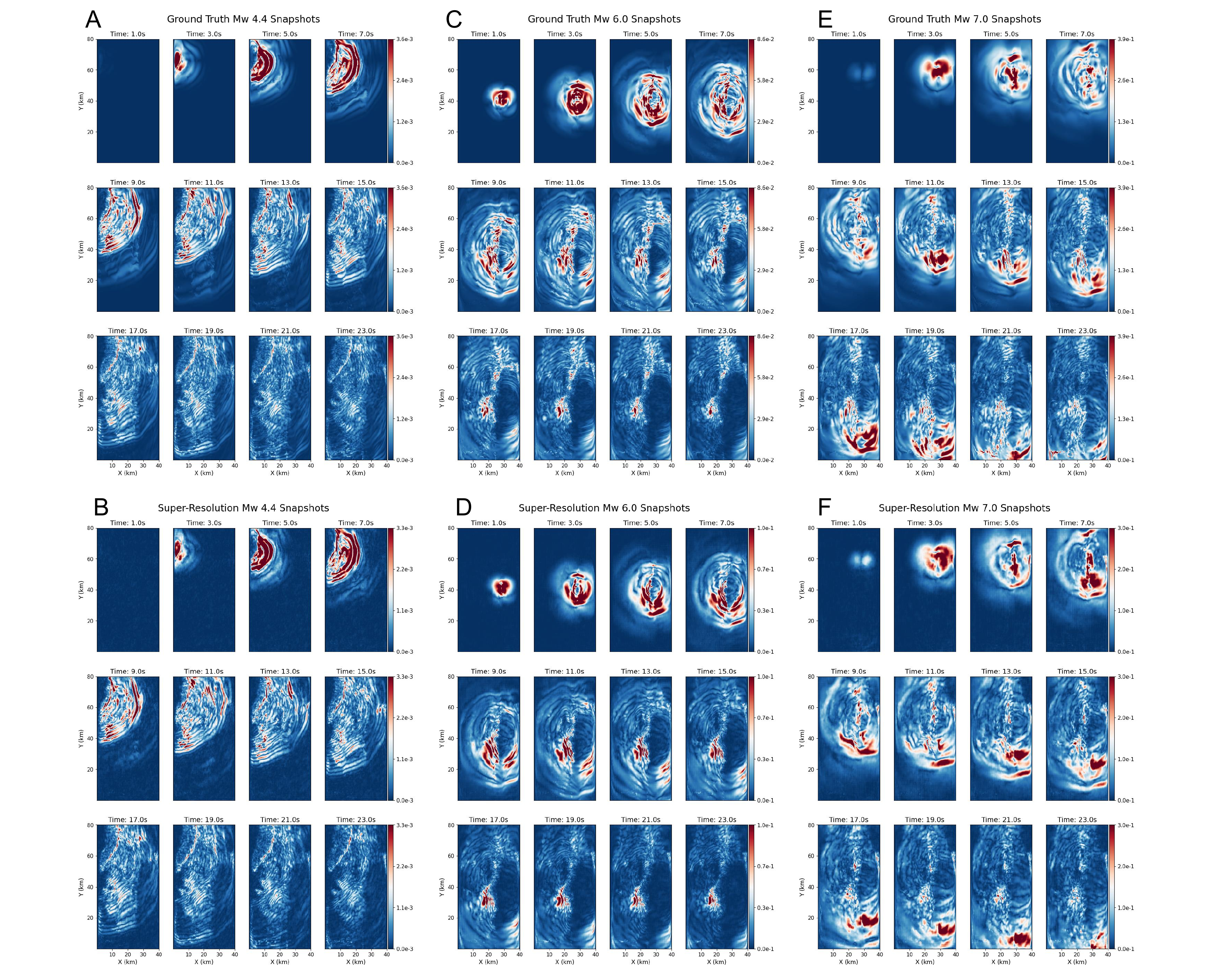}
    \caption{\textbf{Zero-shot super-resolution.} Ground-truth wavefield snapshots are provided at spatial resolution $256 \times 128$, whereas the synthetic samples are super-resolved to spatial resolution $512 \times 256$. \textbf{(A, C, E)} Ground-truth snapshots for representative $M_w\,4.4$, $M_w\,6$, and $M_w\,7$ events, respectively. \textbf{(B, D, F)} Corresponding zero-shot super-resolved synthetic snapshots for the same events.}
\label{fig:sup_res_supp}   
\end{figure*}

\clearpage
\newpage
\section{Ablation Study}

In this section, we perform an ablation study focusing on two factors: 
(i) the capacity of the latent flow matching model (16M vs.\ 128M parameters), and 
(ii) the cutoff frequency used in the \textit{Physics-Aligned Subspace}.

\paragraph{Configuration of flow matching.} We choose the modified 3D diffusion-UNet architecture as the backbone for the flow matching~\cite{tong_improving_2024}.
For the smaller 16M model, we use the following configuration:
\begin{verbatim}
model_config = {
    'hidden_channels': 64,
    'num_heads': 4,
    'attention_res': '8',
}
\end{verbatim}

For the larger 128M model, we use:
\begin{verbatim}
model_config = {
    'hidden_channels': 96,
    'num_res_blocks': 2,
    'num_heads': 8,
    'attention_res': '8,4',
    'channel_mult': (1, 2, 3, 4)
}
\end{verbatim}

\paragraph{Evaluation metrics.}
We reproduce both the statistical distribution analysis and the residual analysis. 
For the statistical distribution comparison, we use the 1D Wasserstein distance. 
Given two one-dimensional probability mass functions \(p_1\) and \(p_2\), the first Wasserstein distance is defined as
\begin{equation}
    l_1(p_1, p_2) = \inf_{\pi \in \Pi(p_1, p_2)} \int_{\mathbb{R}\times\mathbb{R}} |y_1 - y_2| \, d\pi(y_1, y_2),
\end{equation}
where \(\Pi(p_1, p_2)\) denotes the set of all joint distributions on \(\mathbb{R}\times\mathbb{R}\) whose marginals are \(p_1\) and \(p_2\), respectively. 
We quantify distributional agreement by repeating the same procedure used in Fig.~\ref{fig:scaling}A--B. For the residual analysis, we compare the mean residual curves using the root mean squared error (RMSE) as the evaluation metric.

As shown in Table~\ref{tab:ablation_cfm}, lower values indicate better agreement with the target statistics. Increasing model capacity consistently improves performance across nearly all metrics, with the clearest gains observed for the more challenging finite-rupture cases. Likewise, Table~\ref{tab:ablation_fc} suggests that higher cutoff frequencies, particularly \(f_c = 0.75\) and \(1.0~\mathrm{Hz}\), are more effective for capturing large finite-rupture events. This trend is further corroborated by the full residual spectra for different \(f_c\) values shown in Fig.~\ref{fig:sup_res_supp}.
\begin{table}[h]

\caption{\textbf{Model-size ablation for latent flow matching.} Quantitative comparison of the 16M and 128M CFM models using PGV, FAS-based Wasserstein distance at multiple frequencies, and RMSE of the mean residual for \(M_w\,4.4\), \(M_w\,6.0\), and \(M_w\,7.0\) events. The lower the better. }

\centering
\renewcommand{\arraystretch}{0.9}
 \footnotesize 
\setlength{\tabcolsep}{3.pt} %
{
\begin{tabular}
{@{}c  !{\vrule width 0.5pt}  c !{\vrule width 0.5pt}ccccccc@{}}
\toprule
Datasets & Model $\downarrow$ Metric $\rightarrow$ & PGV  & FAS-0.25Hz & FAS-0.5Hz & FAS-0.75Hz & FAS-0.96Hz  & residual mean\\ 
\midrule
\multirow{2}{*}{$\text{Mw}~4.4$}
&CFM - 16M  & $7.2 \cdot 10^{-2}$ & $2.3 \cdot 10^{-2}$ & $5.7 \cdot 10^{-2}$ & $1.2 \cdot 10^{-1}$ & $1.0 \cdot 10^{-1}$ & $1.6 \cdot 10^{-1}$ \\
&CFM - 128M & $5.2 \cdot 10^{-2}$ & $1.6 \cdot 10^{-2}$ & $3.9 \cdot 10^{-2}$ & $1.0 \cdot 10^{-1}$ & $8.3 \cdot 10^{-2}$ & $1.1 \cdot 10^{-1}$ \\
\midrule
\multirow{2}{*}{$\text{Mw}~6.0$}
&CFM - 16M  & $6.5 \cdot 10^{-2}$ & $1.2 \cdot 10^{-2}$ & $9.9 \cdot 10^{-2}$ & $2.7 \cdot 10^{-1}$ & $3.2 \cdot 10^{-1}$ & $4.4 \cdot 10^{-1}$ \\
&CFM - 128M & $3.9 \cdot 10^{-2}$ & $2.0 \cdot 10^{-2}$ & $5.8 \cdot 10^{-2}$ & $2.1 \cdot 10^{-1}$ & $2.7 \cdot 10^{-1}$ & $3.4 \cdot 10^{-1}$ \\
\midrule
\multirow{2}{*}{$\text{Mw}~7.0$}
&CFM - 16M  & $1.4 \cdot 10^{-1}$ & $7.6 \cdot 10^{-2}$ & $1.9 \cdot 10^{-1}$ & $3.8 \cdot 10^{-1}$ & $4.7 \cdot 10^{-1}$ & $6.4 \cdot 10^{-1}$ \\
&CFM - 128M & $8.5 \cdot 10^{-2}$ & $2.5 \cdot 10^{-2}$ & $1.3 \cdot 10^{-1}$ & $3.2 \cdot 10^{-1}$ & $4.3 \cdot 10^{-1}$ & $5.6 \cdot 10^{-1}$ \\
\bottomrule
\end{tabular}}
\label{tab:ablation_cfm}
\end{table}

\begin{table}[h]

\caption{\textbf{Cutoff-frequency ablation for the Physics-Aligned Subspace.} Quantitative comparison of models using different cutoff frequencies \(f_c \in \{0.5, 0.6, 0.75, 1.0\}\,\mathrm{Hz}\) across representative \(M_w\,4.4\), \(M_w\,6.0\), and \(M_w\,7.0\) events. Metrics include PGV, FAS-based Wasserstein distance at multiple frequencies, and RMSE of the mean residual. The best and second-best results in each column are highlighted in red and brown, respectively.}

\centering
\renewcommand{\arraystretch}{0.9}
 \footnotesize 
\setlength{\tabcolsep}{3.pt} %
{
\begin{tabular}
{@{}c  !{\vrule width 0.5pt}  c !{\vrule width 0.5pt}ccccccc@{}}
\toprule
Datasets & Model $\downarrow$ Metric $\rightarrow$ & PGV  & FAS-0.25Hz & FAS-0.5Hz & FAS-0.75Hz & FAS-0.96Hz  & residual mean\\ 
\midrule
\multirow{4}{*}{$\text{Mw}~4.4$}
& fc=0.5hz  & $\sbest{5.7 \cdot 10^{-2}}$ & $2.1 \cdot 10^{-2}$ & $\sbest{4.0 \cdot 10^{-2}}$ & $\best{9.6 \cdot 10^{-2}}$ & $\best{5.9 \cdot 10^{-2}}$ & $\best{1.0 \cdot 10^{-1}}$ \\
& fc=0.6hz  & $6.2 \cdot 10^{-2}$ & $2.4 \cdot 10^{-2}$ & $4.8 \cdot 10^{-2}$ & $1.1 \cdot 10^{-1}$ & $\sbest{7.5 \cdot 10^{-2}}$ & $1.2 \cdot 10^{-1}$ \\
& fc=0.75hz & $\best{5.2 \cdot 10^{-2}}$ & $\sbest{1.6 \cdot 10^{-2}}$ & $\best{3.9 \cdot 10^{-2}}$ & $1.0 \cdot 10^{-1}$ & $8.3 \cdot 10^{-2}$ & $\sbest{1.1 \cdot 10^{-1}}$ \\
& fc=1.0hz  & $5.9 \cdot 10^{-2}$ & $\best{1.5 \cdot 10^{-2}}$ & $4.1 \cdot 10^{-2}$ & $\sbest{9.7 \cdot 10^{-2}}$ & $1.1 \cdot 10^{-1}$ & $1.3 \cdot 10^{-1}$ \\
\midrule
\multirow{4}{*}{$\text{Mw}~6.0$}
& fc=0.5hz  & $\best{5.7 \cdot 10^{-2}}$ & $\best{5.7 \cdot 10^{-3}}$ & $\sbest{8.8 \cdot 10^{-2}}$ & $2.7 \cdot 10^{-1}$ & $\best{1.8 \cdot 10^{-1}}$ & $3.5 \cdot 10^{-1}$\\
& fc=0.6hz  & $7.6 \cdot 10^{-2}$ & $7.8 \cdot 10^{-3}$ & $1.2 \cdot 10^{-1}$ & $2.8 \cdot 10^{-1}$ & $2.9 \cdot 10^{-1}$ & $4.1 \cdot 10^{-1}$ \\
& fc=0.75hz & $\best{3.9 \cdot 10^{-2}}$ & $2.0 \cdot 10^{-2}$ & $5.8 \cdot 10^{-2}$ & $\best{2.1 \cdot 10^{-1}}$ & $\sbest{2.7\cdot 10^{-1}}$ & $\sbest{3.4 \cdot 10^{-1}}$ \\
& fc=1.0hz  & $\sbest{5.0 \cdot 10^{-2}}$ & $\sbest{9.9 \cdot 10^{-3}}$ & $\best{7.7 \cdot 10^{-2}}$ & $\sbest{2.1 \cdot 10^{-1}}$ & $2.8 \cdot 10^{-1}$ & $\best{3.0 \cdot 10^{-1}}$ \\
\midrule
\multirow{4}{*}{$\text{Mw}~7.0$}
& fc=0.5hz  & $1.3 \cdot 10^{-1}$ & $7.6 \cdot 10^{-2}$ & $2.0 \cdot 10^{-1}$ & $4.2 \cdot 10^{-1}$ & $2.8 \cdot 10^{-1}$ & $5.8 \cdot 10^{-1}$ \\
& fc=0.6hz  & $1.4 \cdot 10^{-1}$ & $9.2 \cdot 10^{-2}$ & $2.2 \cdot 10^{-1}$ & $3.9 \cdot 10^{-1}$ & $\best{4.0 \cdot 10^{-1}}$ & $6.3 \cdot 10^{-1}$ \\
& fc=0.75hz & $\sbest{8.5 \cdot 10^{-2}}$ & $\best{2.5 \cdot 10^{-2}}$ & $\sbest{1.3 \cdot 10^{-1}}$ & $\sbest{3.2 \cdot 10^{-1}}$ & $\sbest{4.2 \cdot 10^{-1}}$ & $\sbest{5.6 \cdot 10^{-1}}$ \\
& fc=1.0hz  & $\best{8.2 \cdot 10^{-2}}$ & $\sbest{3.1 \cdot 10^{-2}}$ & $\best{1.2 \cdot 10^{-1}}$ & $\best{2.9 \cdot 10^{-1}}$ & $4.3 \cdot 10^{-1}$ & $\best{5.2 \cdot 10^{-1}}$ \\
\bottomrule
\end{tabular}}
\label{tab:ablation_fc}
\end{table}

\begin{figure*}[ht]
    \vspace*{-.3cm}
    \centering

    \includegraphics[width=1\textwidth,trim={8cm, 0, 8cm, 0},clip]
    {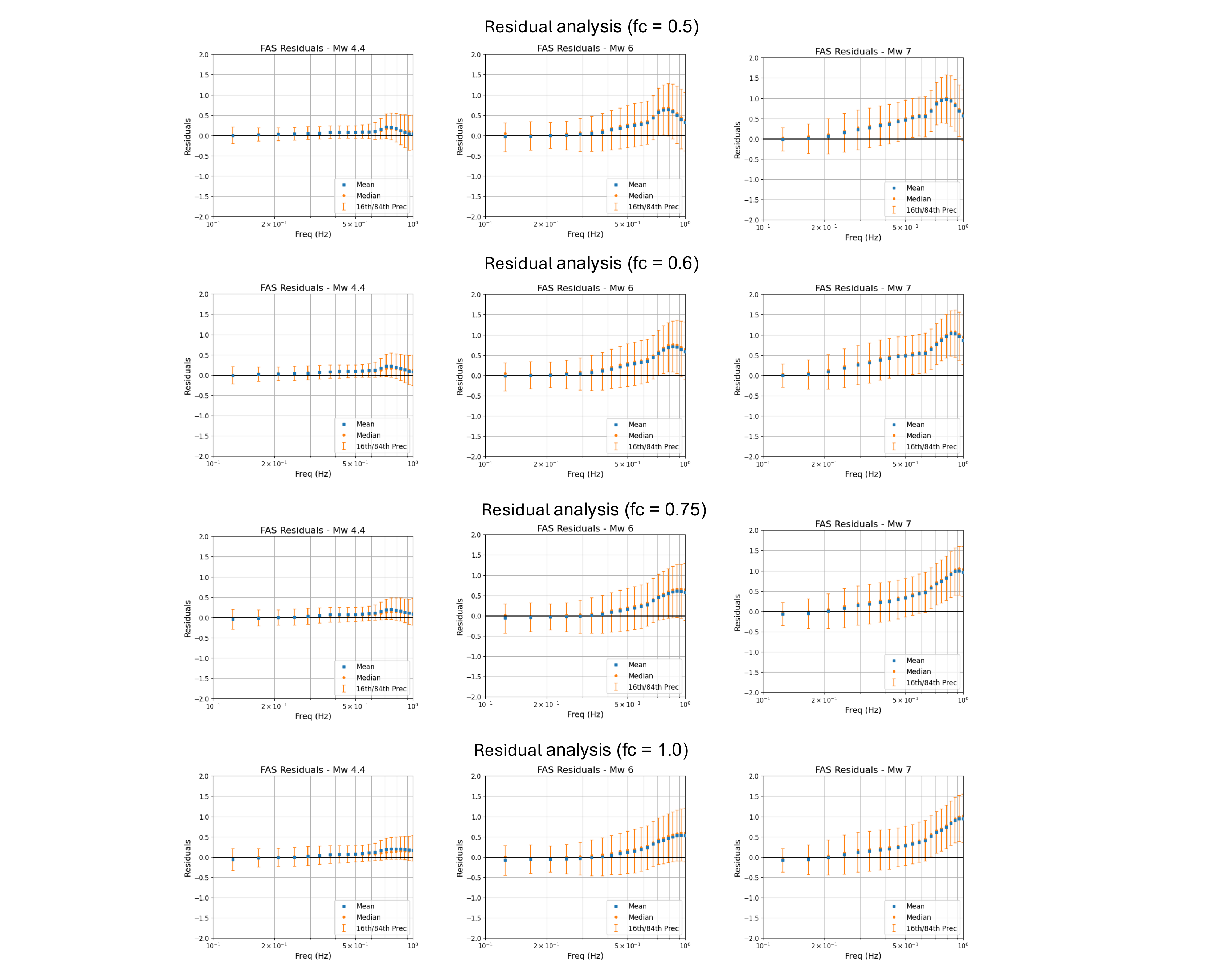}
    \caption{Frequency-dependent spectral residuals for different cutoff frequency $f_c$}  \label{fig:sup_res_supp}   
\end{figure*}

\subsection{Spectral Bias Calibration}

The systematic underestimation of high-frequency content for large finite-rupture events is likely due, at least in part, to limited training coverage in this regime. Our model is trained on only $5{,}000$ samples, which appears sufficient for point-source events, whose wavefields are governed by comparatively simpler propagation effects. In contrast, large-magnitude finite ruptures exhibit richer and more heterogeneous high-frequency structure due to rupture complexity, including extended-source radiation and directivity, and accurately capturing this variability likely requires substantially more training samples.

To mitigate this limitation within the current framework, we introduce an optional post-processing bias correction based on the mean spectral residual shown in Fig.~\ref{fig:residual_analysis}. Empirically, the residual is approximately spatially uniform, suggesting that a single frequency-dependent amplitude correction can be applied to each synthetic event. In practice, the calibration is defined in log-spectral space and applied equivalently as a frequency-dependent multiplicative scaling of the spectrum. This adjustment yields an approximately unbiased spectrum in expectation, as illustrated in Fig.~\ref{fig:residual_analysis_unbiased}. Because the correction modifies amplitudes only, it preserves the predicted phase structure and spatiotemporal coherence while improving FAS agreement and magnitude scaling. Importantly, this correction is only a practical mitigation and does not replace the need for improved generative modeling of high-frequency rupture and scattering effects.

For the representative magnitudes \(M_w\,4.4\), \(6.0\), and \(7.0\), we estimate three corresponding bias functions from the mean spectral residual in log space. To correct a synthetic realization with magnitude between \(M_w\,4.4\) and \(6.0\), or between \(6.0\) and \(7.0\), we linearly interpolate between the two neighboring bias functions to obtain a magnitude-dependent correction curve. This interpolated bias function is then applied as a frequency-dependent multiplicative scaling to the synthetic spectrum.

\begin{figure*}[ht]
    \vspace*{-.3cm}
    \centering

    \includegraphics[width=0.95\textwidth,trim={1cm, 0, 1cm, 0},clip]
    {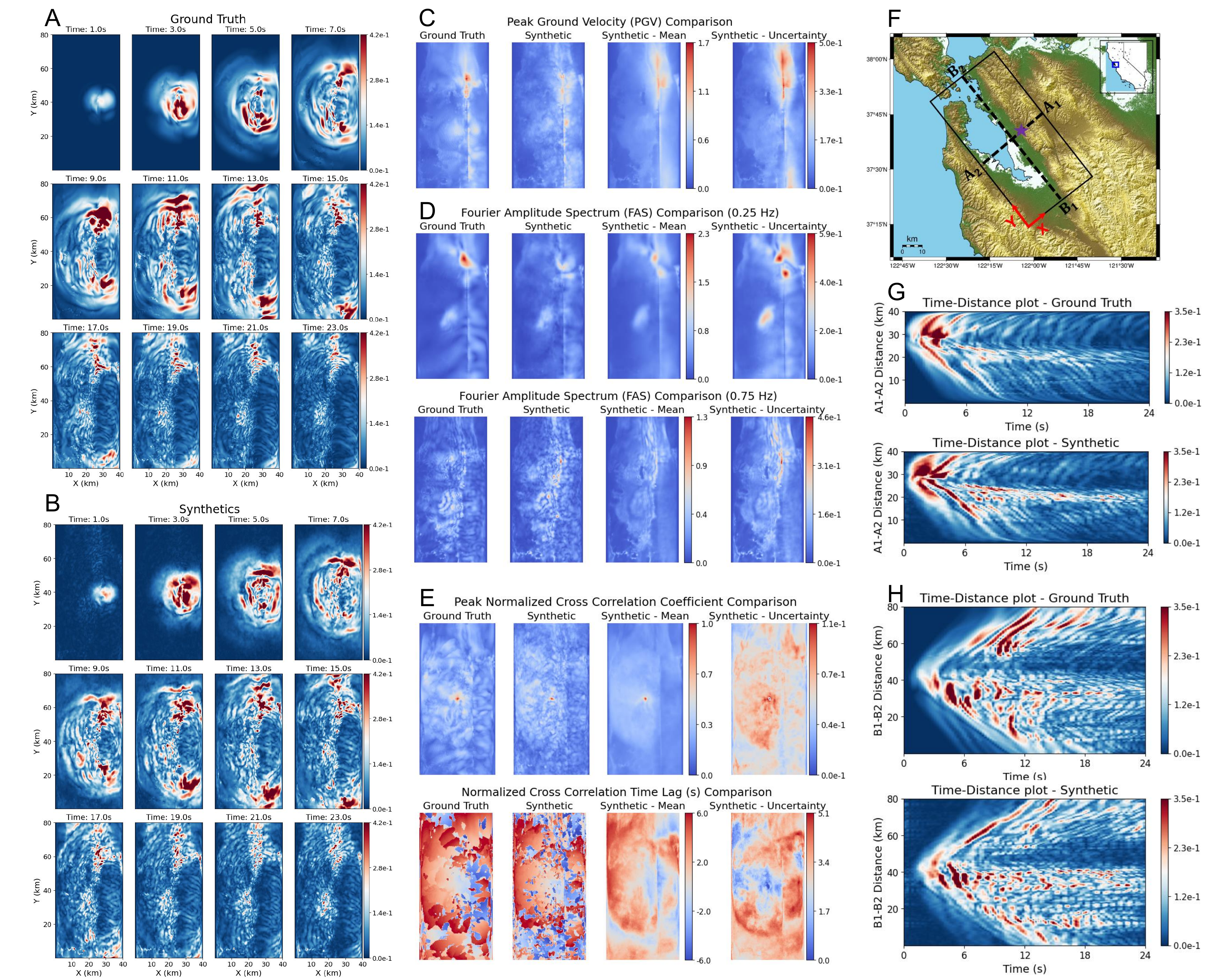}
    \caption{Evaluation of spectral-calibrated \alg on a representative $M_w\,7.0$ point-source scenario.} \label{fig:supp_bias_m7_scen}   
\end{figure*}

Additional results for this calibration procedure are provided in this section. As shown in Table~\ref{tab:ablation_bias}, the calibrated synthetic samples substantially reduce spectral bias relative to the uncalibrated results. As further illustrated in Fig.~\ref{fig:supp_bias_m7_scen}, the calibration does not alter the phase information, and the corrected synthetics remain both visually and physically plausible. Moreover, Fig.~\ref{fig:mag_scaling_bias} shows that, after calibration, the magnitude scaling is much better aligned with the ground truth. Overall, these results confirm that the proposed spectral calibration effectively reduces systematic amplitude bias while preserving the predicted phase structure and spatiotemporal coherence.

\begin{figure*}[ht]
    \vspace*{-.3cm}
    \centering

    \includegraphics[width=0.95\textwidth,trim={1cm, 0, 1cm, 0},clip]
    {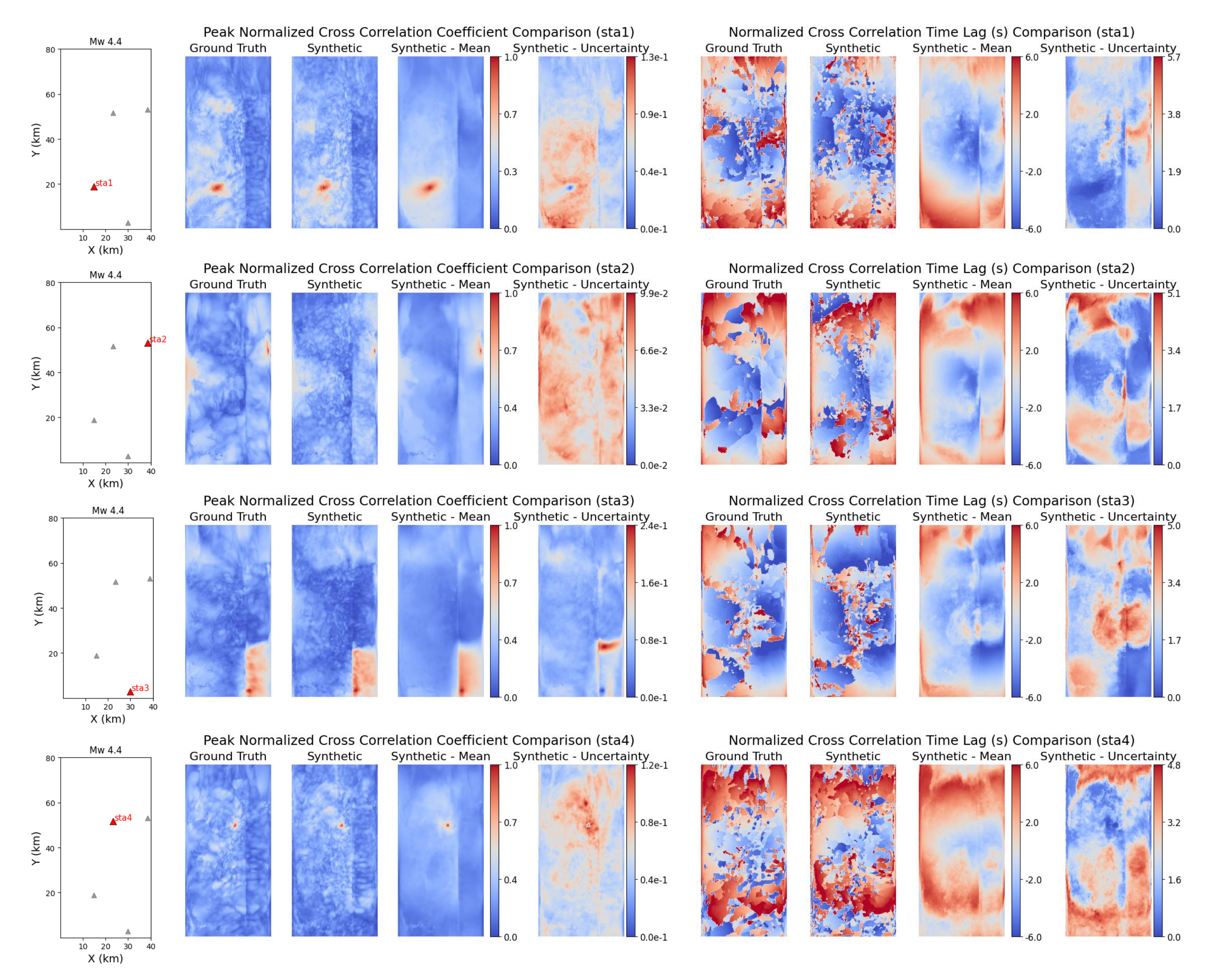}
\caption{NCC analysis using randomly selected reference stations for the spectral-calibrated representative $M_w~7$ event.}
  \label{fig:supp_fig_ncc_bias}   
\end{figure*}

\begin{figure*}[ht]
    \vspace*{-.3cm}
    \centering

    \includegraphics[width=0.85\textwidth,trim={1cm, 0, 1cm, 0},clip]
    {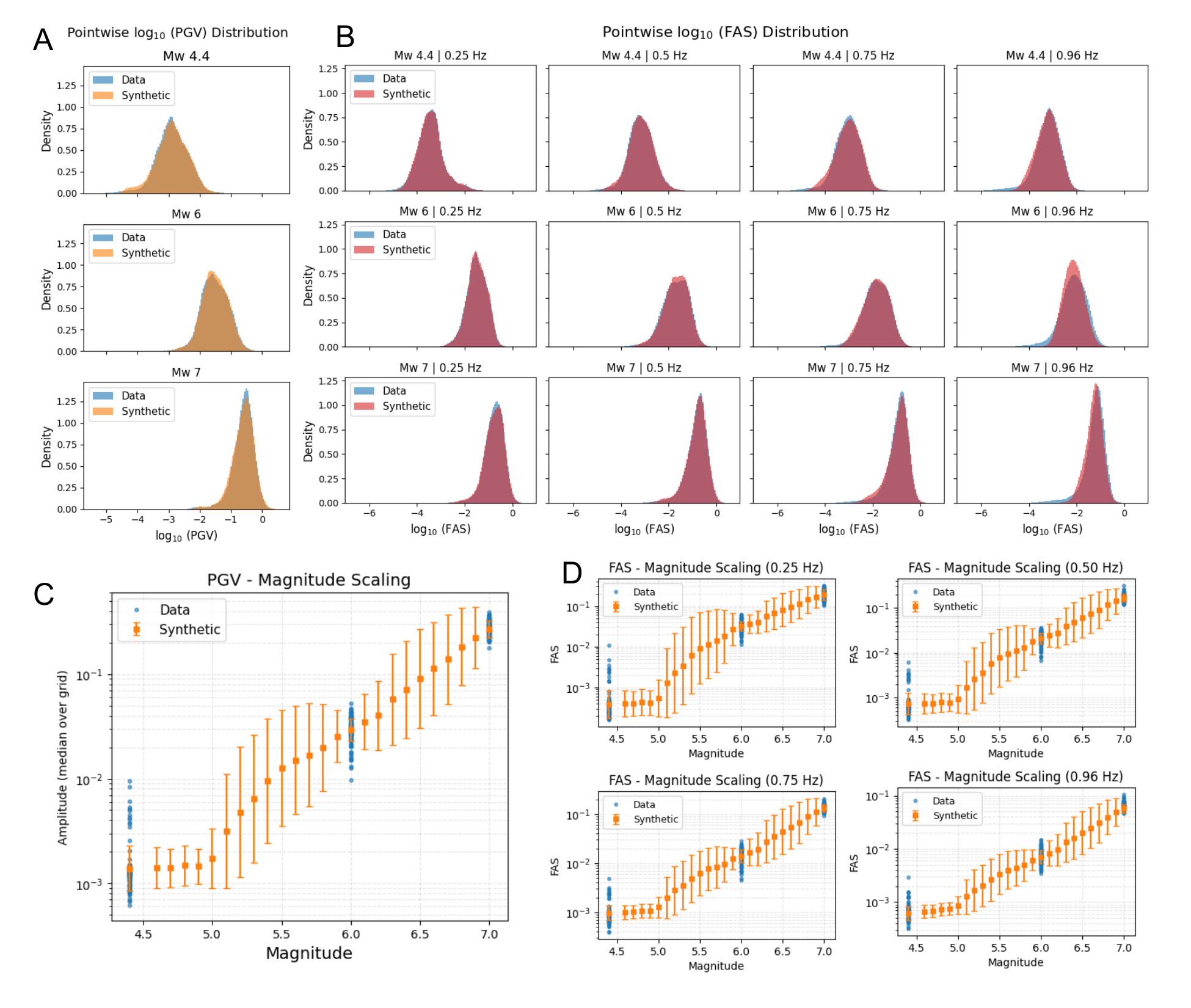}
    \caption{Statistical distribution and magnitude scaling laws of spectral-calibrated generated ground motions}\label{fig:mag_scaling_bias}   
\end{figure*}

\begin{figure*}[ht]
    \vspace*{-.3cm}
    \centering

    \includegraphics[width=0.85\textwidth,trim={1cm, 0, 1cm, 0},clip]
    {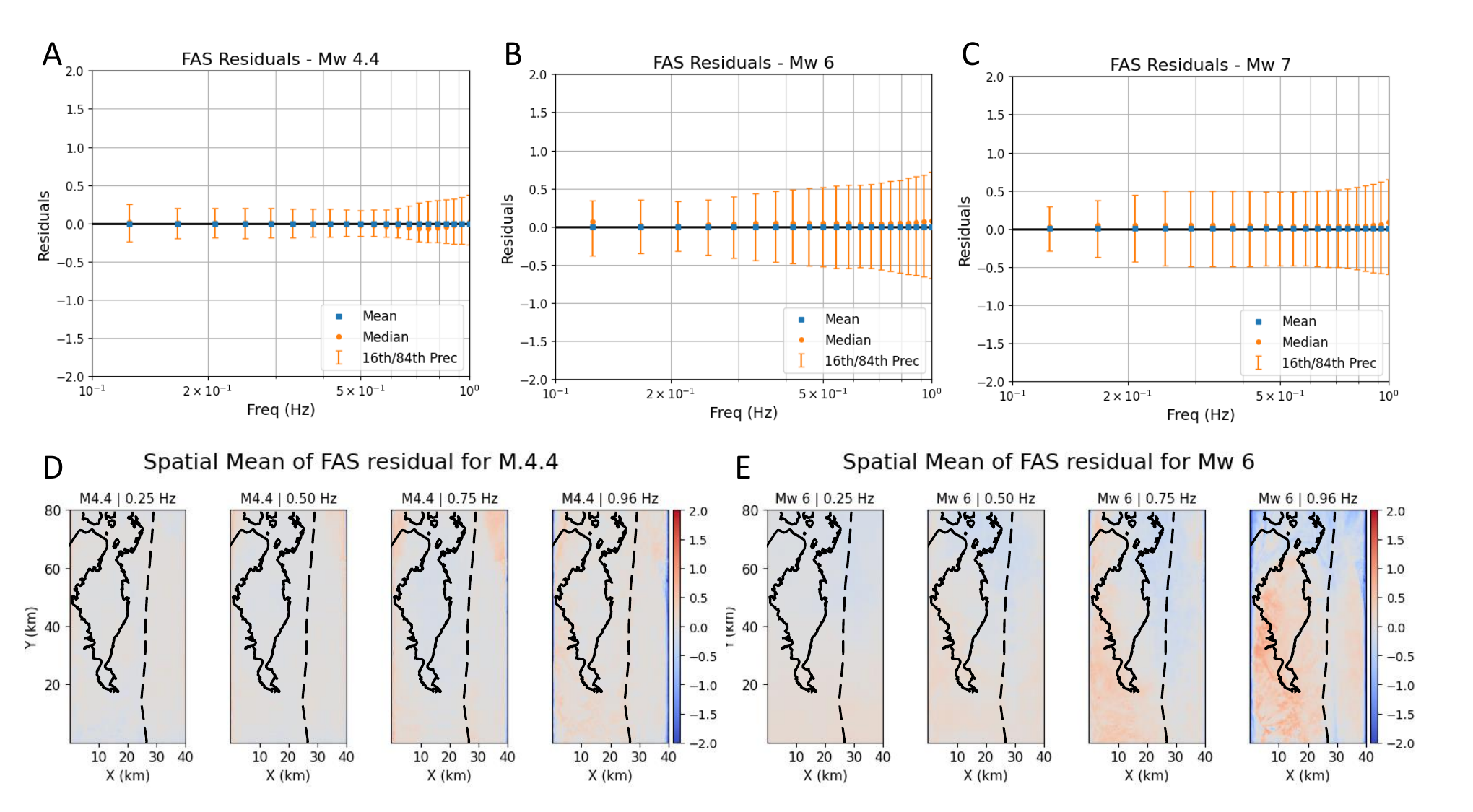}
\caption{Frequency-dependent spectral residual with calibrated synthetics.}
\label{fig:residual_analysis_unbiased}   
\end{figure*}

\begin{table}[h]

\caption{\textbf{Effect of post-processing spectral calibration.} Quantitative comparison of \alg\ and \alg-calibrated for representative \(M_w\,4.4\), \(M_w\,6.0\), and \(M_w\,7.0\) events using PGV, FAS-based Wasserstein distance at selected frequencies, and RMSE of the mean residual. The lower, the better.}

\centering
\renewcommand{\arraystretch}{0.9}
 \footnotesize 
\setlength{\tabcolsep}{3.pt} %
{
\begin{tabular}
{@{}c  !{\vrule width 0.5pt}  c !{\vrule width 0.5pt}ccccccc@{}}
\toprule
Datasets & Model $\downarrow$ Metric $\rightarrow$ & PGV  & FAS-0.25Hz & FAS-0.5Hz & FAS-0.75Hz & FAS-0.96Hz  & residual mean\\ 
\midrule
\multirow{2}{*}{$\text{Mw}~4.4$}
&\alg-calibrated  & $2.1 \cdot 10^{-2}$ & $1.2 \cdot 10^{-2}$ & $1.0 \cdot 10^{-2}$ & $2.9 \cdot 10^{-2}$ & $4.6 \cdot 10^{-2}$ & $5.0 \cdot 10^{-4}$ \\
& \alg & $5.2 \cdot 10^{-2}$ & $1.6 \cdot 10^{-2}$ & $3.9 \cdot 10^{-2}$ & $1.0 \cdot 10^{-1}$ & $8.3 \cdot 10^{-2}$ & $1.1 \cdot 10^{-1}$ \\
\midrule
\multirow{2}{*}{$\text{Mw}~6.0$}
&\alg-calibrated   & $1.4 \cdot 10^{-2}$ & $9.6 \cdot 10^{-3}$ & $2.3 \cdot 10^{-2}$ & $1.6 \cdot 10^{-2}$ & $9.8 \cdot 10^{-2}$ & $1.2 \cdot 10^{-3}$ \\
&\alg  & $3.9 \cdot 10^{-2}$ & $2.0 \cdot 10^{-2}$ & $5.8 \cdot 10^{-2}$ & $2.1 \cdot 10^{-1}$ & $2.7 \cdot 10^{-1}$ & $3.4 \cdot 10^{-1}$ \\
\midrule
\multirow{2}{*}{$\text{Mw}~7.0$}
&\alg-calibrated   & $1.8 \cdot 10^{-2}$ & $1.5 \cdot 10^{-2}$ & $1.2 \cdot 10^{-2}$ & $2.8 \cdot 10^{-2}$ & $7.8 \cdot 10^{-2}$ & $2.9 \cdot 10^{-2}$ \\
&\alg & $8.5 \cdot 10^{-2}$ & $2.5 \cdot 10^{-2}$ & $1.3 \cdot 10^{-1}$ & $3.2 \cdot 10^{-1}$ & $4.3 \cdot 10^{-1}$ & $5.6 \cdot 10^{-1}$ \\
\bottomrule
\end{tabular}}
\label{tab:ablation_bias}
\end{table}

\subsection{Legend for movie files}
The movie files compare generated ground motions with the corresponding simulation data. \texttt{Movie 1}, \texttt{Movie 2}, and \texttt{Movie 3} show data-versus-synthetic comparisons for the $M_w~4.4$, $M_w~6$, and $M_w~7$ scenarios, respectively. \texttt{Movies 4, 5, and 6} show data-versus-spectrally calibrated synthetic comparisons for the $M_w~4.4$, $M_w~6$, and $M_w~7$ scenarios, respectively.

\end{document}